\documentclass[dvipsnames]{article} %
\usepackage{colm2024_conference}

\usepackage{booktabs}
\usepackage{graphicx}
\usepackage{enumitem}
\usepackage{wrapfig}
\usepackage{algorithm}
\usepackage{algpseudocode}

\usepackage{microtype}
\usepackage{amsmath}
\usepackage{colortbl}
\usepackage[utf8]{inputenc}
\definecolor{lightgray}{rgb}{0.9,0.9,0.9}
\usepackage{caption}
\usepackage{subcaption}
\usepackage{setspace}
\usepackage{url}
\usepackage{multirow}
\usepackage{colortbl}
\usepackage{tabularx}
\usepackage{blindtext}
\usepackage{pgfplots}
\pgfplotsset{compat=1.18}
\usepackage{tikz}
\usetikzlibrary{er,positioning,bayesnet}
\usepackage{makecell}
\usepackage{tipa}
\usepackage{siunitx}
\usepackage{nicefrac}
\usepackage{tocloft}
\usepackage{listings}
\usepackage[raster,skins]{tcolorbox} %
\usepackage{xltabular}
\usepackage{adjustbox}
\usepackage{xurl}
\usepackage{rotating}
\usepackage[normalem]{ulem}
\usepackage{kotex} 
\useunder{\uline}{\ul}{}

\usepackage{amsmath,amsfonts,bm}

\def\eqref#1{equation~\ref{#1}}

\def\1{\bm{1}}

\DeclareMathAlphabet{\mathsfit}{\encodingdefault}{\sfdefault}{m}{sl}
\SetMathAlphabet{\mathsfit}{bold}{\encodingdefault}{\sfdefault}{bx}{n}

\newcommand*\justify{%
  \fontdimen2\font=0.4em
  \fontdimen3\font=0.2em
  \fontdimen4\font=0.1em
  \fontdimen7\font=0.1em
  \hyphenchar\font=`\-
}

\renewcommand{\texttt}[1]{%
  \begingroup
  \ttfamily
  \begingroup\lccode`~=`/\lowercase{\endgroup\def~}{/\discretionary{}{}{}}%
  \begingroup\lccode`~=`[\lowercase{\endgroup\def~}{[\discretionary{}{}{}}%
  \begingroup\lccode`~=`.\lowercase{\endgroup\def~}{.\discretionary{}{}{}}%
  \catcode`/=\active\catcode`[=\active\catcode`.=\active
  \justify\scantokens{#1\noexpand}%
  \endgroup
}

\title{A.X K2 Technical Report}

\author{
\bf SK Telecom
}

\begin{document}

\maketitle

\begin{abstract}
We introduce A.X K2, a 688B-parameter Mixture-of-Experts (MoE) language model trained from scratch as a high-performance foundation for \emph{agentic} applications. Trained on approximately 8.5T tokens---fewer than its predecessor, A.X K1---on a smaller but higher-quality mixture with substantially expanded agentic and software-engineering data, it nonetheless improves over A.X K1 across the board, by over 30 percentage points on some benchmarks, reflecting large gains in token efficiency. To support long contexts efficiently, we introduce Sparse Gated Attention (SGA), which combines sparse attention with gated attention, and adopt Gated Norm (GN) to stabilize large-scale training. SGA is trained natively at 128K through a \emph{sparse} indexer warmup that optimizes the indexer against its own sparse top-$k$ selection rather than the dense attention distribution, making adaptation markedly cheaper: each query reads only 2,048 positions, yet long-context quality is unchanged and A.X K2 scores 94.6 on RULER out to 256K. The outlier suppression of GN in turn keeps 4-bit NVFP4 serving within one point of FP8 accuracy. A simple yet effective Think-Fusion recipe further lets users switch between thinking and non-thinking modes within a single unified model. Extensive evaluations show that A.X K2 performs competitively against strong open-weight baselines, matching or exceeding them on math and Korean-language benchmarks.
\end{abstract}

\begin{figure}[!b]
    \centering
    \includegraphics[width=\linewidth]{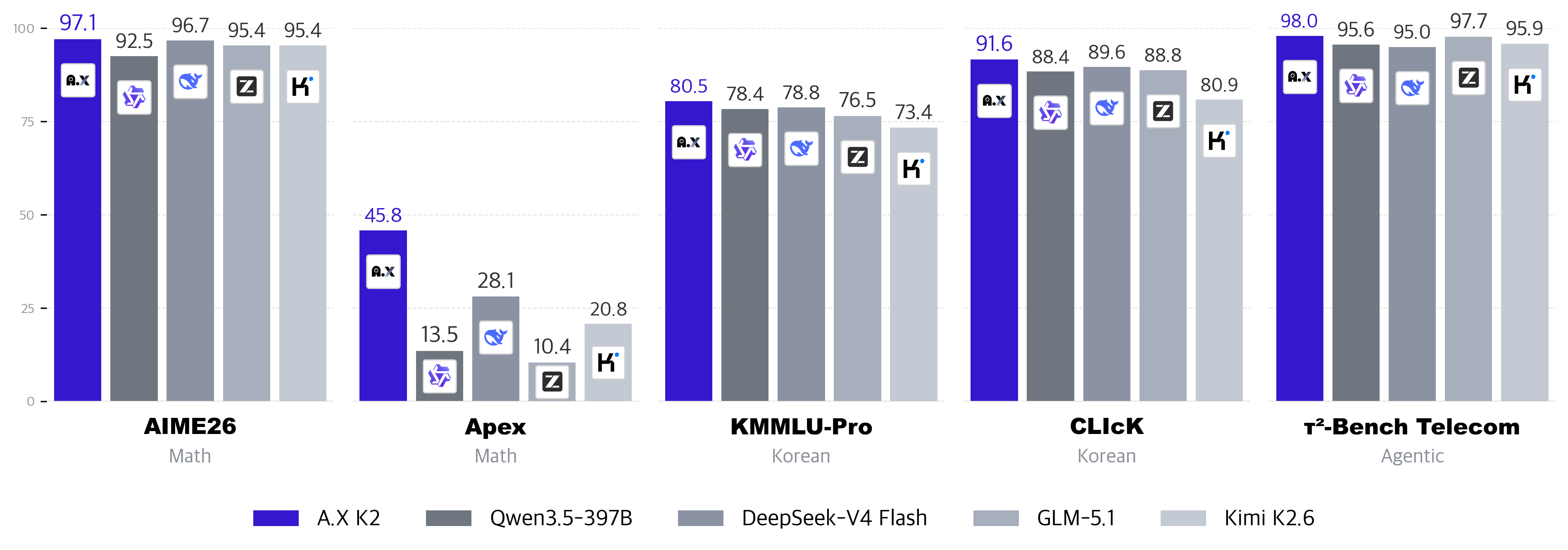}
    \caption{\textbf{Performance comparison of A.X K2 against leading open-weight models across representative benchmarks} (thinking mode; cf.\ Table~\ref{tab:LLM_large_latest_think}). A.X K2 leads on mathematics (AIME26, Apex), Korean-language understanding (KMMLU-Pro, CLIcK), and agentic tool use ($\tau^2$-Bench Telecom). } 
    \label{fig:teaser}
\end{figure}

\section{Introduction}

Large language models (LLMs) have advanced rapidly through the scaling of model capacity and advances in reinforcement learning, and the frontier is increasingly defined by \emph{agentic} competence---multi-step reasoning, tool use, and long-horizon task execution~\citep{deepseekai2026deepseekv4, kimiteam2025kimik2openagentic, qwen2026qwen35omni, glm5team2026glm5}. Turning such capabilities into deployable systems, however, poses challenges that raw scale alone does not resolve: a practical foundation model must pair broad knowledge and strong reasoning with efficient serving, controllable inference-time compute, and trainability under realistic data and hardware budgets.

We present \textbf{A.X K2}, a 688B-parameter Mixture-of-Experts (MoE) language model with 33B active parameters, trained from scratch as a high-performance, agentic foundation model and the successor to A.X K1~\citep{skt2026axk1tr}. Guided by MoE scaling laws~\citep{tian2025greaterleveragescalinglaws}, the architecture maximizes knowledge capacity within fixed hardware constraints while prioritizing inference throughput over strictly compute-optimal training. A.X K2 is trained on approximately 8.5T tokens in total---about 8.2T in pre-training and the remainder in post-training, fewer than A.X K1---using a smaller but higher-quality mixture oriented toward agentic workloads, yet it remains competitive with leading open-weight models, reflecting substantial gains in token efficiency.

Because deep reasoning is costly---long chains of thought inflate latency and serving cost~\citep{snell2024scaling} and are wasteful on simple queries---A.X K2 gives users explicit control over the extent of reasoning. Through a simple yet effective \textit{Think-Fusion} recipe, a single unified model supports both a \emph{thinking} mode for complex problems and a \emph{non-thinking} mode for concise, low-latency responses, allowing deployments to trade quality for cost on a per-request basis.

Beyond its technical contributions, A.X K2 is part of a national \textit{Sovereign AI} effort. By building a frontier-scale model that deeply understands the Korean language and culture, we aim to reduce reliance on foreign proprietary systems and accelerate the adoption of trustworthy, controllable AI across Korean industry, government, and academia.

We highlight four key aspects of A.X K2 as follows:

\begin{itemize}
    \item \textbf{Token-Efficient Pre-training:} We demonstrate large-scale MoE
    training with 688B total parameters (and 33B active parameters) using
    approximately 8.5T tokens in total ($\sim$8.2T in pre-training), guided by scaling laws under a fixed compute budget.
    Despite training on fewer tokens than A.X K1~\citep{skt2026axk1tr} ($\sim$10T), A.X K2 shows
    substantial improvements across the board---over 30 percentage points on some benchmarks---indicating
    improved token efficiency driven by an
    enhanced multi-stage data-processing pipeline, with further headroom available
    from additional data.

    \item \textbf{Gated Transformer Blocks for Stable Training and Efficient Long-Context Inference:} We introduce \emph{Gated Transformer Blocks}, which integrate \textbf{Sparse Gated Attention (SGA)} and \textbf{Gated Norm (GN)}. SGA combines a lightweight indexer~\citep{deepseekai2025deepseekv32} with gated attention~\citep{qiu2025gatedattention}: the sparse attention prunes the attention computation to curb long-context compute---each query attends to only 2,048 positions, or 1.6\% of a 128K context---while the head-specific output gate adds non-linearity and suppresses attention sinks, improving loss convergence and attention quality. We further contribute a \emph{sparse warmup} recipe for the adaptation stage: rather than first fitting the indexer to the dense (full) attention distribution before enabling sparse selection, we optimize the indexer directly over its own sparse top-$k$ selection from the outset, which makes the warmup substantially cheaper at negligible quality cost (Sec.~\ref{subsec:long_context}). The resulting sparsity is quality-neutral---LongBench is essentially unchanged across the adaptation (62.80 $\rightarrow$ 62.99)---while long-context serving throughput and latency improve markedly (Sec.~\ref{sec:evaluation}). GN~\citep{qiu2026unified} further mitigates outliers in the hidden states---massive activations that inflate per-block scales---while improving the accuracy of low-bit formats such as FP8 and FP4. Together, these designs let A.X K2 extend its usable context length while remaining amenable to low-precision deployment, reducing memory footprint and inference latency on long-context workloads.

    \item \textbf{Think-Fusion Supervised Fine-Tuning (SFT) and Multi-Stage Reinforcement Learning:} Our \textit{Think-Fusion} SFT recipe trains a single unified model on paired thinking and non-thinking responses, so that mode switching is learned from explicit control tokens rather than from superficial distributional cues. The resulting model supports a \textit{non-thinking mode} for concise, low-latency responses and a \textit{thinking mode} for extensive reasoning chains, empowering users to flexibly allocate compute at inference time based on task complexity. Subsequent multi-stage reinforcement learning jointly optimizes instruction following, human-preference alignment, agentic tool use, and safety under a shared reward framework.

    \item \textbf{Frontier-Scale Engineering Capability:} We demonstrate the feasibility of end-to-end engineering for a 688B-parameter model, including stable large-scale training, reproducible pipelines, and system-level optimizations, further establishing a foundation for scaling toward frontier-class model development.
\end{itemize}

Extensive evaluations show that A.X K2 performs competitively against strong open-weight baselines across English and Korean benchmarks---notably matching or exceeding them on math and Korean tasks---while providing explicit, user-controllable switching between thinking and non-thinking modes at inference time. Beyond benchmarks, our results validate the feasibility of unifying implicit and explicit reasoning processes within a single parameter space. We release A.X K2\footnote{\url{https://huggingface.co/skt/A.X-K2}}, positioning it as a practical, controllable foundation model.

\paragraph{Organization.} The remainder of this report details the model architecture and tokenizer design (Sec.~\ref{sec:architecture}), pre-training (Sec.~\ref{sec:pretrain}), post-training (Sec.~\ref{sec:posttrain}), and evaluation results (Sec.~\ref{sec:evaluation}).

\section{Architecture}
\label{sec:architecture}

\begin{figure*}[t!]
    \centering
    \includegraphics[width=\textwidth]{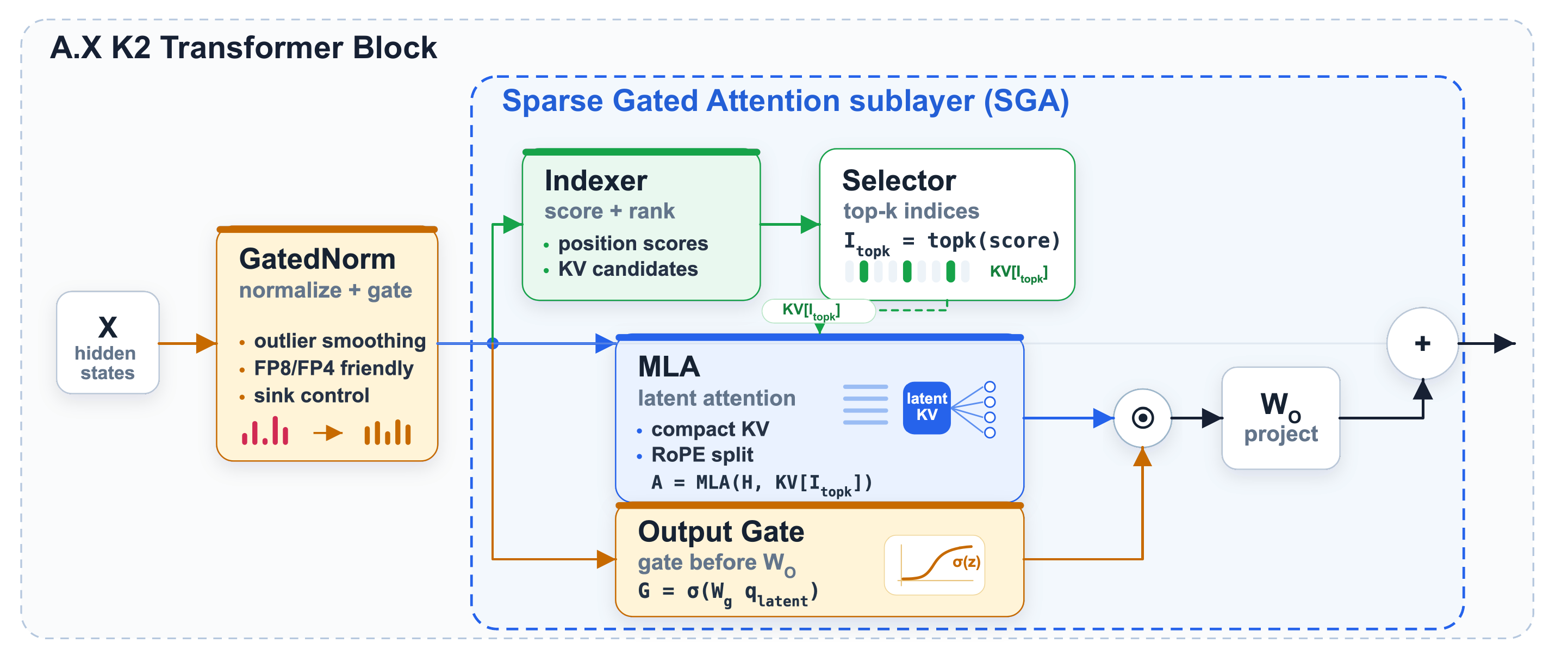}
    \caption{\textbf{The A.X K2 transformer block with the SGA sublayer.} A lightweight indexer scores and ranks key--value candidates, and the selector keeps the top-$k$ indices $I_{\text{topk}}$; MLA then performs latent attention over the selected KV cache $\mathrm{KV}[I_{\text{topk}}]$. A head-specific output gate $G=\sigma(W_g\,q_{\text{latent}})$ modulates the attention output before the output projection $W_O$. GN precedes the sublayer, smoothing outliers for stable, low-precision (FP8/FP4) computation.}
    \label{fig:sga}
\end{figure*}

\begin{table}[htbp]
\footnotesize
\centering
\begin{adjustbox}{max width=\textwidth,center}
\begin{tabular}{@{}lcccccccccc@{}}
\toprule
Models
& \makecell{Total \\ Params.}
& \makecell{Activated \\ Params.}
& Layers
& \makecell{Heads \\ (Q / KV)}
& \makecell{Hidden \\ Size}
& \makecell{Intermediate Size \\ (Dense / Expert)}
& \makecell{Routed / Activated \\ Experts}
& \makecell{Shared \\ Experts}
& \makecell{Vocab \\ Size}
& \makecell{Context Length \\ (Native / YaRN)} \\
\midrule
A.X K2       & 688B & 33B  & 61 & 64 & 7168 & 18432 / 2048 & 256 / 8 & 1 & 163,840 & 128K / 256K \\
\bottomrule
\end{tabular}
\end{adjustbox}
\caption{\textbf{Architecture overview of A.X K2, an MoE model built from A.X K2 transformer blocks.} The 128K context is trained natively via ABF, and 256K is reached at inference by increasing the YaRN scaling factor (Sec.~\ref{subsec:long_context}).}
\label{tab:ax}
\end{table}

\subsection{Model Configuration}
The A.X K2 model is a large-scale MoE language model with 688B total parameters and 33B active parameters, developed as part of the Korean government's `Sovereign AI foundation model' project.\footnote{The Ministry of Science and ICT (MSIT), Republic of Korea, through the National IT Industry Promotion Agency (NIPA) (Grant No.\ PJT-26-010018).} Detailed architectural specifications are presented in Table~\ref{tab:ax}. For the pre-training compute budget, given a training timeline of approximately 70 days on 512 NVIDIA B200 GPUs, we estimated the total available training FLOPs as a fixed compute budget, under which architectural trade-offs were guided by established scaling principles for MoE models.

Relative to A.X K1, the principal change in model scale is an increase in the number of routed experts from 192 to 256, which raises total capacity while keeping the activated parameter count at 33B. The expert count was set to meet the target total-parameter budget derived from the available FLOPs, and we chose 256---a power of two and a multiple of 128---to align with expert-parallel sharding and kernel tiling for implementation robustness.

The architectural design of A.X K2 adopts several widely used components that have proven effective in large-scale MoE systems~\citep{liu2024deepseek, yang2025qwen3technicalreport, 5team2025glm45agenticreasoningcoding}. The backbone uses Multi-head Latent Attention (MLA)~\citep{liu2024deepseek}, which improves Key--Value cache efficiency and reduces memory overhead under long-context settings. On top of MLA, we apply a head-specific output gate (gated attention)~\citep{qiu2025gatedattention} throughout pre-training; the gate introduces non-linearity into the attention output, mitigates attention sinks, and improves loss convergence. For additional training stability, the query and key projections are normalized before the attention scores are computed (QK-normalization)~\citep{dehghani2023scaling}. Of the 61 transformer layers, the first is a dense feed-forward layer and the remaining 60 are MoE layers; the expert-routing and load-balancing strategy is detailed in Section~\ref{sec:pretrain}.

To improve training stability and mitigate hidden-state outliers, we adopt Gated Norm (GN)~\citep{qiu2026unified}, which applies a gating operation immediately after RMSNorm~\citep{zhang2019rootmeansquarelayer}. By suppressing outlier amplification in the normalized output, GN yields smoother loss convergence. As an additional benefit, it produces an activation distribution with suppressed outliers that is favorable for low-precision (e.g., NVFP4) serving. The gating operation of GN incurs an approximately 5\% reduction in training throughput, a cost we accepted in exchange for these stability and low-precision benefits; with GN in place, training remained sufficiently stable \emph{without} the dual-normalization scheme used in A.X K1.

Whereas A.X K1 stabilized training with a dual-normalization scheme---inspired by Gemma~\citep{team2025gemma} and early GLM-4 designs that stack normalization layers around the attention and MLP blocks---A.X K2 removes this redundancy entirely. The single post-normalization gating of GN subsumes the stabilizing effect of the earlier dual-normalization design while reducing architectural complexity.

Separately, informed by empirical observations in \citep{kimiteam2025kimik2openagentic}, we configured the model with 64 attention heads and integrated shared dense experts as an operating point that prioritizes inference efficiency by reducing attention overhead while enhancing knowledge sharing.

The tokenizer is inherited unchanged from A.X K1: a byte-level BPE~\citep{sennrich2016neural} vocabulary of 163,840 tokens covering English, Korean, Chinese, Japanese, and Spanish.

\subsection{Gated Transformer Blocks for Mitigating Attention Sinks and Outliers}
\label{subsec:gated_blocks}

\begin{wrapfigure}{r}{0.5\textwidth}
	{
	\includegraphics[width=75mm]{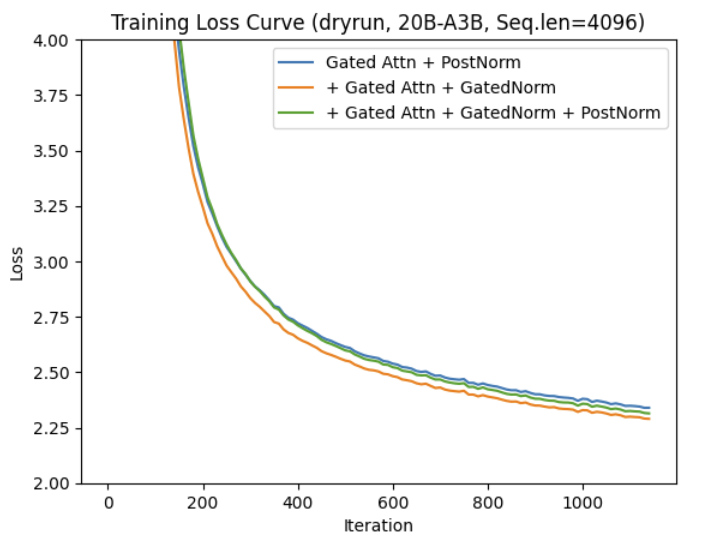}
    \caption{\textbf{Training loss for the GN ablation (dry-run, 20B-A3B, sequence length 4096).} Adding GN on top of gated attention improves loss convergence, while further stacking a post-MLP normalization layer yields no additional benefit, indicating that GN alone subsumes the stabilizing effect of the earlier dual-normalization design.}
    \label{fig:ablation-gated}
	}
\end{wrapfigure}

\paragraph{Sparse Gated Attention.}
To make long-context inference efficient while keeping training stable, A.X K2 combines two mechanisms into \textbf{SGA} (Figure~\ref{fig:sga})\footnote{We use ``SGA'' for our design; despite the shared name, it is unrelated to the method of~\citet{du2025sgamf}.}: gated attention~\citep{qiu2025gatedattention}, applied throughout pre-training as described above, and a sparse-attention mechanism added for long contexts. For sparsity, we adopt the sparse-attention mechanism of~\citet{deepseekai2025deepseekv32}, which uses a lightweight indexer to select a small set of top-$k$ tokens ($k=2048$) per query, substantially reducing attention compute at long sequence lengths. The budget is fixed rather than proportional, so its relative cost shrinks as the context grows: at a 128K context each query attends to 1.6\% of the available positions and at 256K to 0.8\%, making the attention term scale essentially linearly rather than quadratically in sequence length. As described in Sec.~\ref{sec:pretrain}, this sparse component is introduced through a dedicated adaptation stage, in which the indexer is trained with a \emph{sparse warmup and adaptation} after the model has been natively trained to a 128K context, so that long-context quality is preserved while inference cost is reduced. The gating, by contrast, is not specific to long context: it is the head-specific output gate already present in every attention layer during pre-training. The two mechanisms are also mutually reinforcing: because the indexer is trained to reproduce the attention distribution, its top-$k$ quality is bounded by the quality of that distribution. By suppressing the attention-sink mass that standard softmax attention places on a few uninformative tokens~\citep{qiu2025gatedattention}, the output gate yields a better-calibrated signal so that the indexer spends its limited budget on genuinely relevant positions. Crucially, introducing the sparse component leaves long-context quality essentially intact: on LongBench~\citep{bai2024longbench}, A.X K2 scores 62.80 before and 62.99 after SGA adaptation (Table~\ref{tab:sga-longbench}), so the efficiency gains below are obtained at no measurable quality cost. In long-context serving, these properties translate into concrete efficiency gains: as the input length grows into the tens of thousands of tokens, A.X K2 with SGA sustains substantially higher total- and output-token throughput than A.X K1~\citep{skt2026axk1tr}, as shown in Figure~\ref{fig:throughput-k1-k2} and analyzed in detail in Sec.~\ref{sec:inference_optimization}.

\paragraph{Gated Norm.}

A.X K2 additionally adopts \textbf{GN}, in which a learned, input-dependent gate modulates the normalized activation before it enters the residual stream. Beyond its stability benefits, GN bounds the magnitude of activations propagated across layers: it suppresses \emph{massive activations}, a small number of hidden units that are orders of magnitude larger than the rest and persist across layers, which are a well-documented source of both training instability and low-precision quantization error~\citep{qiu2025gatedattention, qiu2026unified}. This matters for deployment under narrow block-scaled formats such as NVFP4, where 4-bit (E2M1) elements share a single scale over a small block (e.g., 16 elements): a lone outlier inflates that block's scale and collapses the precision of all co-located values, an effect further amplified in the MLA latent KV cache through the up-projection. By keeping the per-channel activation distribution well behaved, GN, together with the SGA output gate, tightens the per-block dynamic range that low-precision formats must absorb, improving block-scale utilization and lowering FP4/W4A4 error.

\section{Pre-Training}
\label{sec:pretrain}

\subsection{Pre-Training Dataset}
A.X K2 is pre-trained on approximately 8.2 trillion tokens spanning diverse web text, code, STEM (Science, Technology, Engineering, and Mathematics), reasoning, books, and synthetic data. The corpus is built with the multi-stage data-processing pipeline developed for A.X K1~\citep{skt2026axk1tr}, which combines an in-house document-parsing pipeline (a Korean-specialized vision-language model with fine-tuned layout detection that recovers text from PDFs), a dual synthetic-data pipeline (seed-corpus-based reasoning and agentic data, and topic-based knowledge synthesis), and a three-stage curation pipeline: heuristic and model-based quality filtering with deduplication, domain-aware mixture and upsampling, and difficulty scoring for curriculum learning~\citep{bengio2009curriculum}. As these components are largely unchanged from A.X K1, we refer the reader to~\citet{skt2026axk1tr} for their details and focus here on the data updates specific to A.X K2.

For A.X K2, we refreshed and expanded this corpus along several axes. The English web and code corpora were updated to more recent snapshots---principally the Nemotron-CC v2.1 corpus~\citep{su2025nemotron, nvidia2025nemotronccv21} and its code counterpart, Nemotron-Pretraining-Code-v2~\citep{nvidia2025nemotroncodev2}---and the Korean web corpus was likewise refreshed, while multilingual coverage was broadened by incorporating FineWeb2~\citep{penedo2025fineweb2}. New synthetic data was generated using stronger open-source models alongside in-house models~\citep{SKTAdotX4}. Reflecting the agentic and long-horizon objectives of A.X K2, we substantially increased the proportion of agentic and software-engineering data, including tool-use trajectories and repository-level coding data, as well as high-quality long-context data drawn from web and PDF sources. In the later stages we additionally blended in instruction-style (SFT-format) data to strengthen instruction-following and task competence ahead of post-training. The resulting stage-wise category mixture is summarized in Appendix~\ref{appendix:pretrain_data} and illustrated in Figure~\ref{fig:curriculum}.

\paragraph{Personal information.} Corpora collected in-house pass an internal PII masking pipeline before training. Detected identifiers---primarily email addresses and numeric identifiers such as phone numbers---are replaced with type-specific placeholder tokens (for example \texttt{<|email|>}) rather than deleted, so that the surrounding text remains intact and the document stays usable as training signal; detection rules are maintained per identifier type and reviewed for false positives. Publicly released datasets are used as distributed and are not re-processed, since their publishers already apply their own PII handling.

\subsection{Quality Filtering and Difficulty-Aware Curation}
Although the overall curation framework is inherited from A.X K1, the filtering regime for A.X K2 was substantially revised to operate under a tighter compute budget. The full candidate pool---roughly 16.2 trillion tokens---far exceeded what could be consumed within the project's compute and schedule constraints, so rather than indiscriminately training on all available data, we prioritized the most \emph{training-efficient} subset and arranged it as a difficulty-increasing curriculum, concentrating the highest-value, highest-difficulty data in the later stages of training~\citep{hu2024minicpm, dubey2024llama, olmo2025olmo3, bengio2009curriculum}. We assess each document along two axes---quality and domain---and progressively tighten both as training advances.

\paragraph{A stage-dependent filtering scheme.} Quality and domain are assessed by progressively stricter criteria across stages, as summarized in Table~\ref{tab:filtering_scheme}. In Stage~1, quality is measured by an \emph{educational-value} score and documents are partitioned by a coarse \emph{web-domain} classifier into 26 general topical categories.\footnote{Arts and entertainment, autos and vehicles, beauty and fitness, books and literature, business and industrial, computers and electronics, finance, food and drink, games, health, hobbies and leisure, home and garden, internet and telecom, jobs and education, law and government, news, online communities, people and society, pets and animals, real estate, science, sensitive subjects, shopping, sports, travel and transportation, and adult content.} From Stage~2 onward, once the corpus has been concentrated on higher-quality data, quality is measured by educational value \emph{and} difficulty, and documents are re-partitioned by a finer \emph{expert-domain} classifier over 14 academic and professional fields.\footnote{Mathematics, physics, chemistry, biology, earth science, manufacturing engineering, other engineering, IT, humanities, social science, law, economics, medicine, and culture.} In both regimes, the quality threshold is set \emph{per domain} rather than globally so that it tracks each domain's distribution, and it is raised in Stage~2.

\begin{table}[t]
\centering
\small
\begin{tabular}{@{}lll@{}}
\toprule
 & Stage~1 & Stage~2 onward \\
\midrule
Quality signal & educational value & educational value $+$ difficulty \\
Domain taxonomy & web domains (26 topical) & expert domains (14 academic) \\
Per-domain threshold & baseline & raised \\
Curricular role & broad coverage & high-difficulty concentration \\
\bottomrule
\end{tabular}
\caption{\textbf{Stage-dependent filtering.} Both the quality signal and the domain taxonomy are refined from Stage~1 to Stage~2, once the corpus has been concentrated on higher-quality data.}
\label{tab:filtering_scheme}
\end{table}

Both scores are produced by a compact A.X-Encoder--based classifier distilled from a larger teacher: the teacher labels a seed set, the classifier is trained on these labels, and inference over the full corpus yields per-document educational-value and difficulty scores on a $0$--$4$ scale. Restricting the difficulty signal to Stage~2 and beyond induces the difficulty-increasing curriculum, and the per-domain thresholds are deliberately relaxed for fields dominated by long-tail knowledge so as to preserve rare content.

\paragraph{Web corpora and stage-wise consumption.} The two principal web sources are an aggregated English web corpus ($\approx$9.1T tokens), of which Nemotron-CC v2.1~\citep{su2025nemotron, nvidia2025nemotronccv21} is the largest single contributor alongside several additional English web sources, and an in-house Korean crawl ($\approx$1.37T tokens). Under the scheme above, the English web supplies $\approx$3.68T unique tokens in Stage~1 and $\approx$0.22T in Stage~2, and the Korean web supplies $\approx$1.12T and $\approx$0.06T, respectively; all per-stage figures are \emph{unique}-token counts, obtained by capping each dataset's epoch multiplicity at one (Table~\ref{tab:web_token_budget}).

\begin{table}[t]
\centering
\small
\begin{tabular}{@{}lrrr@{}}
\toprule
Source & Corpus & Stage~1 & Stage~2 \\
\midrule
English Web (primarily Nemotron-CC v2.1) & 9.1T & 3.68T & 0.22T \\
Korean Web (in-house crawl) & 1.37T & 1.12T & 0.06T \\
\bottomrule
\end{tabular}
\caption{\textbf{Web-corpus sizes and the unique-token volume consumed per stage.} All counts are measured with the A.X K2 tokenizer (Sec.~\ref{sec:architecture}) and therefore need not match the token counts published with the source datasets, which use different tokenizers.}
\label{tab:web_token_budget}
\end{table}

\subsection{Pre-Training Process}

\begin{figure*}[t!]
    \centering
    \includegraphics[width=\textwidth]{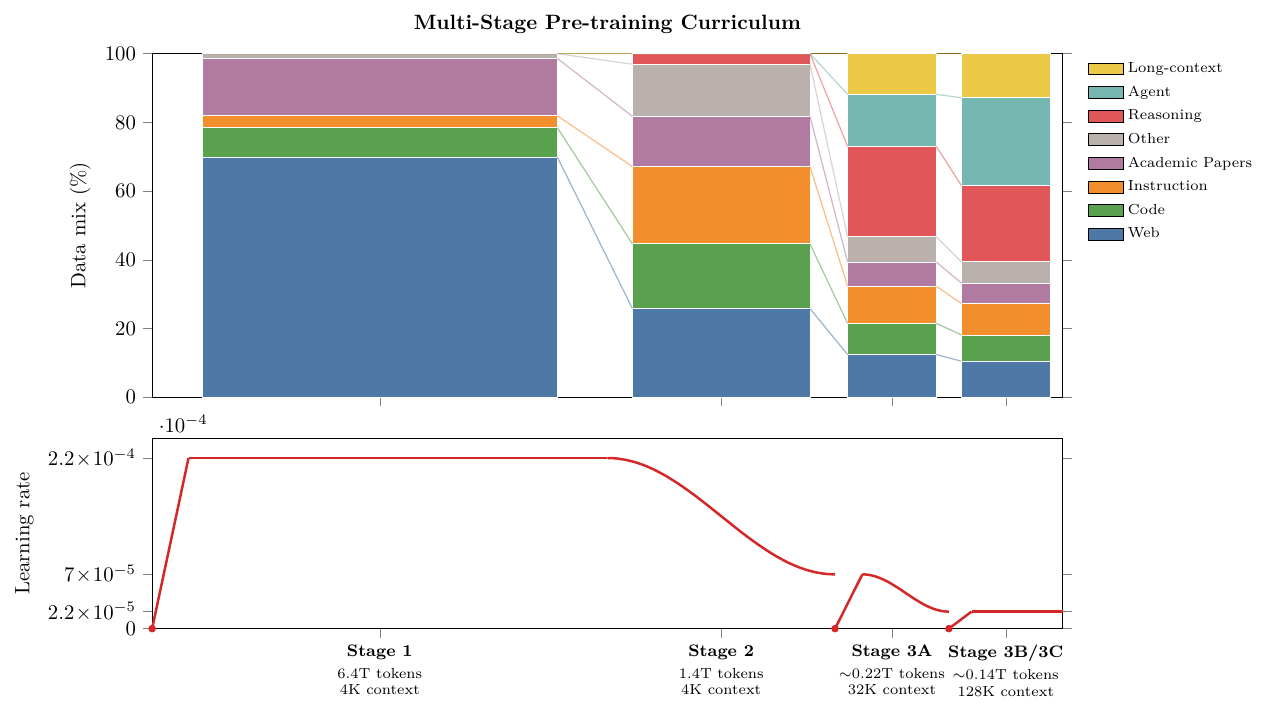}
    \caption{\textbf{Multi-stage pre-training curriculum of A.X K2.} \textbf{Top:} the stage-wise shift in data-mixture composition (\%); earlier-introduced sources sit at the bottom of each bar and newly-introduced data (reasoning, agent, long-context) on top, with light connectors tracing each category across stages. \textbf{Bottom:} the Warmup--Stable--Decay (WSD) learning-rate schedule, where each of the three warm-ups (Stage~1, 3A, 3B) rises from zero and is followed by a cosine decay. Token counts and context lengths are annotated per stage; stage widths are illustrative and not to token scale.}
    \label{fig:curriculum}
\end{figure*}

To effectively optimize A.X K2 across varying data distributions, we adopt a Warmup-Stable-Decay (WSD) learning-rate schedule~\citep{hu2024minicpm} across a multi-stage pre-training pipeline (Figure~\ref{fig:curriculum}). Pre-training is organized into three stages, defined by their training objectives, data characteristics, and corresponding phase of the WSD schedule: \emph{Stage~1} (general knowledge), \emph{Stage~2} (advanced knowledge and reasoning), and \emph{Stage~3} (long-context extension). Stages~1 and~2 are described below, while Stage~3 is detailed separately in Section~\ref{subsec:long_context}.

\begin{enumerate}
    \item \textit{Stage~1 --- General Knowledge (Warmup and Stable Phase):} A.X K2 is trained on a corpus of 6.4 trillion tokens at a sequence length of 4,096, establishing a robust foundation in both linguistic proficiency and general world knowledge. After an initial learning-rate warmup and batch-size ramp-up, the model is trained at the constant maximum learning rate for the remainder of the stage---the stable phase of the WSD schedule---to maximize knowledge acquisition.

    \item \textit{Stage~2 --- High-Quality Reasoning (Decay Phase):} Using a curated corpus of 1.4 trillion tokens at a sequence length of 4,096, this stage is strictly filtered for high-complexity web data and further incorporates a diverse mixture of domain-specific datasets, including STEM, reasoning-intensive tasks, refined PDF content, and high-quality synthetic data. Here we begin the learning-rate decay, annealing from the maximum learning rate to stabilize the model on these high-quality distributions.

    \item \textit{Stage~3 --- Long-Context Adaptation:} We progressively extend the usable context length to 32K and then 128K, followed by a sparse-attention adaptation. Owing to its distinct objectives and recipe, this stage is detailed separately in Section~\ref{subsec:long_context}.
\end{enumerate}

\subsection{Long-Context Adaptation}
\label{subsec:long_context}

Stage~3 enables long-context capabilities through three sub-stages: we natively extend the context length to 32K (Stage~3A, approximately 0.22T tokens) and subsequently to 128K (Stage~3B), followed by sparse-attention adaptation (Stage~3C; described below), rather than relying on post-hoc interpolation. Stages~3B and~3C together consume approximately 0.14T tokens. For the extension, we follow the Adjusted Base Frequency (ABF) approach~\citep{xiong2023effective}: the RoPE base is kept at its default value of $10^{4}$ throughout Stages~1 and~2 (sequence length 4,096), raised to $10^{6}$ at the start of the 32K extension (Stage~3A), and then held at $10^{6}$ for the 128K extension (Stage~3B) and the subsequent SGA adaptation (Stage~3C), so that the full 128K context is learned natively without any further base adjustment. In our experiments, this native base adjustment yielded stronger long-context retrieval than relying on RoPE interpolation alone. This design is motivated by the demands of agentic tasks, which require long-horizon reasoning, planning, and action over extended contexts; accordingly, the long-context stages incorporate data featuring long chains of thought and repository-level code.

\begin{figure*}[t!]
    \centering
    \includegraphics[width=\textwidth]{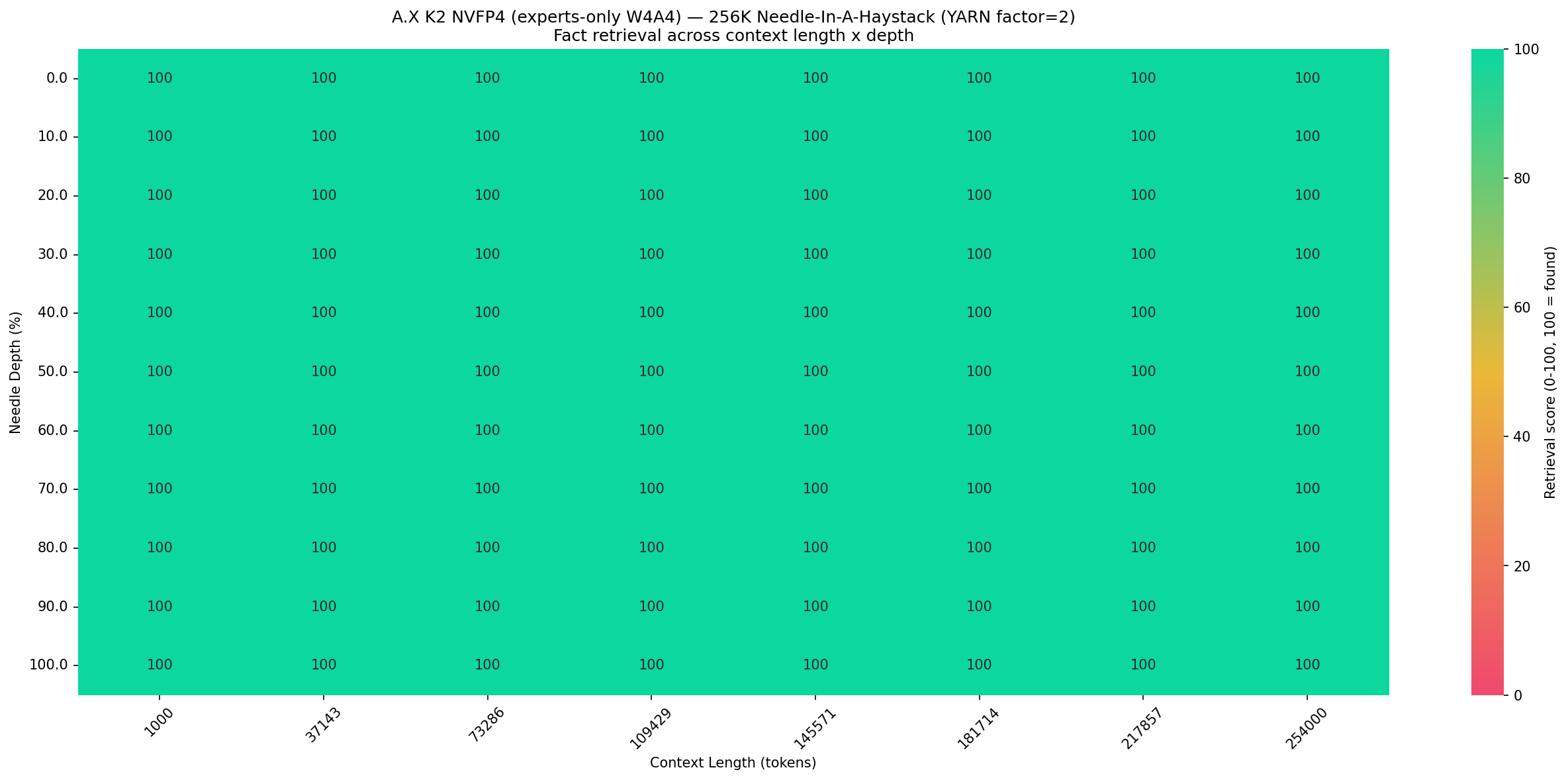}
    \caption{\textbf{Needle-in-a-haystack (NIAH) retrieval for A.X K2 at a 256K context length, obtained by increasing the YaRN scaling factor.} Each cell reports the retrieval score (0--100, 100 = found) across context length (x-axis) and needle depth (y-axis); A.X K2 attains a perfect score of 100 at every position, including under NVFP4 experts-only W4A4 quantization.}
    \label{fig:niah_256k}
\end{figure*}

For RoPE scaling, we use YaRN~\citep{peng2023yarn}, which can be applied in two modes. A \emph{static} mode uses a single fixed scaling factor across the entire sequence, which favors fast convergence during training (as adopted by DeepSeek and Kimi-K2~\citep{liu2024deepseek, kimiteam2025kimik2openagentic}); a \emph{dynamic} mode instead leaves positions within the originally trained length unscaled and interpolates only beyond it, recomputing the factor at inference time from the actual sequence length to minimize extrapolation error (as adopted by Qwen~\citep{yang2025qwen3technicalreport}). We keep the YaRN attention-temperature scaling factor (\emph{mscale}) at 1.0, applying no additional rescaling to the attention logits. Because A.X K2 reaches 128K natively through ABF, longer contexts are supported by increasing the YaRN scaling factor, with the static or dynamic mode chosen by the serving stack. With an enlarged scaling factor---and with the RoPE base itself left unchanged at $10^{6}$---A.X K2 extends to a 256K context and attains strong needle-in-a-haystack (NIAH) retrieval at this length, as shown in Fig.~\ref{fig:niah_256k}.

Finally, after reaching 128K, we apply a dedicated adaptation stage (Stage~3C) that introduces the sparse-attention component of SGA (Sec.~\ref{subsec:gated_blocks}) to improve long-context inference efficiency. Stage~3C reuses the same data mixture as Stage~3B---the two share a mixture, which differs from that of Stage~3A---so that introducing SGA does not require switching to a separate dataset, following the approach proposed in GLM-5~\citep{glm5team2026glm5}. The SGA indexer, which selects the top 2{,}048 tokens per query, is trained by a KL-divergence loss that aligns its selection scores with the attention distribution. We begin Stage~3C with an indexer-only warmup, freezing all non-indexer parameters so that the indexer adapts without perturbing the already-converged backbone. Crucially, unlike DeepSeek-V3.2~\citep{deepseekai2025deepseekv32} and GLM-5~\citep{glm5team2026glm5}, which first warm up the indexer against the \emph{dense} (full) attention distribution before enabling sparse selection, we compute the indexer loss directly over the sparse top-$k$ selection from the outset---i.e., a \emph{sparse} warmup. After this warmup, the full model resumes training with the indexer loss down-weighted, producing the final sparse-attention model. Because the warmup runs under sparse attention rather than full dense attention, it is substantially faster than a dense warmup; in our experiments, this yielded a large speedup while incurring negligible difference in downstream performance, making sparse warmup the more cost-effective choice.

\subsection{Checkpoint Merging}

As the final step of pre-training, we apply checkpoint merging---a Warmup-Stable-Merge (WSM) style of Stochastic Weight Averaging (SWA)~\citep{izmailov2018swa}---to produce the released base model. Rather than committing to a single final checkpoint, we average the parameters of the last six checkpoints, spaced approximately 1.6B tokens apart (25 optimizer steps of $\approx$64M tokens each), directly in weight space. Averaging over nearby points along the optimization trajectory reduces the variance of the final solution and tends to settle in a flatter region of the loss landscape, improving robustness and generalization at no additional training cost. The merge is computed shard-wise over all model tensors, including the MoE router weights, which we found beneficial to average rather than exclude. We use the $1-\sqrt{\cdot}$ weighting scheme, which assigns larger weight to the more recent checkpoints, so that the merged model stays close to the latest, best-converged checkpoint while still benefiting from averaging; uniform and exponential-moving-average weightings are also supported. Because the averaged checkpoints span only a narrow window (roughly 8B tokens in total) and therefore lie close together in parameter space, the merge incurs negligible cost while consistently yielding a stronger final model than any individual checkpoint in the window.

\subsection{Hyperparameters}

The magnitudes of the learning rate and batch size were selected based on the MoE scaling laws proposed by \citet{tian2025greaterleveragescalinglaws}, which were introduced earlier in the context of model size and architecture design. We optimize with AdamW ($\beta_1 = 0.9$, $\beta_2 = 0.95$, $\epsilon = 10^{-8}$), following GPT-3~\citep{gpt3}. The peak learning rate was set to $2.2\times 10^{-4}$ and the global batch size to 16,384 sequences (with a micro-batch size of 4), corresponding to approximately 64M tokens per step at a sequence length of 4,096. Following the WSD schedule, the learning rate is linearly warmed up over the first $\sim$6.1M sequences ($\approx$25B tokens) and then held constant at $2.2\times10^{-4}$ through the general stage; it is subsequently cosine-decayed to $7\times10^{-5}$ across the reasoning stage. Each long-context extension stage then re-warms the learning rate at its start---over roughly 0.3M sequences for the 32K stage and 0.08M sequences for the 128K stage---to stabilize optimization as the sequence length and data distribution change: the 32K stage warms up to $7\times10^{-5}$ and cosine-decays to $2.2\times10^{-5}$, and the 128K stage warms up to and then holds $2.2\times10^{-5}$, a level maintained through the SGA-adaptation stage.

To mitigate gradient spikes and ensure stability during the early training phase, a global batch-size ramp-up was applied, linearly increasing the batch size from 2,048 to 16,384 in increments of 2,048 over the first 72 million sequences (approximately 295B tokens at a sequence length of 4,096). As the sequence length grows during long-context training, the number of sequences per step is reduced accordingly (for example, to 2,048 sequences at a 32K length) so that the tokens processed per step remain close to 64M throughout.

In addition to the general optimization hyperparameters, several MoE-specific settings were configured to balance routing stability and expert utilization. Tokens are routed by a sigmoid-scored, pre-softmax router using grouped top-$k$ selection: the 256 routed experts are partitioned into 8 groups, and each token activates the top 8 experts under a group top-$k$ of 4, with the routing logits moderated by a top-$k$ scaling factor of 2.5. A.X K1 used a router-bias mechanism together with a sequence-level auxiliary loss~\citep{skt2026axk1tr, liu2024deepseek}. In contrast, A.X K2 adopts the global auxiliary loss of~\citet{qiu2025demonsdetailimplementingload}. A sequence-level loss enforces expert balance within each individual sequence---an overly local constraint that suppresses expert specialization and can degrade quality---whereas computing the balancing loss over a larger, global batch relaxes this constraint while still equalizing expert utilization. The global auxiliary loss alone, however, did not sufficiently reduce the maximal violation (MaxVio) of expert load~\citep{wang2024auxiliarylossfreeloadbalancingstrategy}---the worst-case relative deviation of an expert's load from the uniform target---so we additionally retain a learnable expert-bias term (whose update rate adapts across stages), measuring MaxVio over the global batch rather than the per-microbatch quantity used in the original formulation.

The load-balancing strength is annealed as training progresses. During the general and reasoning stages, the (global, sequence-level) auxiliary-loss coefficients are set to $(4\times10^{-3},\,\text{off})$---the sequence-level term is only monitored rather than applied---and the expert-bias update rate is $1\times10^{-3}$ to facilitate adaptation of expert selection. For the long-context extension stages, the coefficients are tightened to $(1\times10^{-3},\,1\times10^{-5})$ and the bias update rate is lowered to $1\times10^{-4}$ to stabilize routing as the data distribution shifts toward higher-quality and long-context data. During the final SGA adaptation stage, the coefficients are further reduced to $(5\times10^{-4},\,2\times10^{-5})$; the expert bias is briefly frozen (update rate $0$) while the sparse-attention indexer---which selects the top 2{,}048 tokens per query---is warmed up, and is then resumed at $1\times10^{-4}$.

\subsection{Parallelism and Training Efficiency}

Pre-training was conducted on 512 NVIDIA B200 GPUs (64 nodes of 8 GPUs) over approximately 70 days. We used pipeline parallelism (PP) of 8 with an interleaved virtual-pipeline degree of 2, and expert parallelism (EP) of 8, with the remaining ranks assigned to data parallelism; tensor parallelism was not used. The larger per-device memory of the B200---relative to the H200 used for A.X K1---allowed us to reduce pipeline parallelism from 16 to 8 while still holding the model without tensor parallelism, thereby lowering pipeline-bubble overhead. We employed a distributed optimizer, which partitions optimizer states across data-parallel ranks and achieves memory savings comparable to ZeRO Stage~2~\citep{rajbhandari2020zeromemoryoptimizationstraining}. For expert parallelism, token dispatch and combine were handled by the HybridEP all-to-all backend rather than DeepEP, with expert-parallel communication confined to the eight-GPU intra-node NVLink domain and a small number of streaming multiprocessors (24--32) dedicated to the dispatch/combine kernels. These settings were chosen on the basis of large-scale profiling to preserve the intended total training compute within the project timeline.

Context parallelism was employed to support long-context training: a context-parallel degree of 2 was used for the 32K extension stage and a degree of 8 for the 128K stage, distributing attention computation across devices while keeping per-GPU memory usage within practical limits.

To manage activation memory pressure, A.X K2 uses \emph{full} activation recomputation~\citep{korthikanti2022reducingactivationrecomputationlarge}, applied uniformly with each transformer layer forming its own recomputation unit. Although full recomputation incurs additional recompute cost relative to the selective scheme used in A.X K1~\citep{skt2026axk1tr}, the memory it frees allowed us to raise the micro-batch size to 4; in our setup this trade-off---full recomputation with a larger micro-batch---delivered higher per-GPU throughput than selective recomputation with a smaller micro-batch, by better amortizing kernel-launch and pipeline overheads across more samples. At the longer sequence lengths of the long-context stages (Stage~3A onward), however, even full recomputation left limited memory headroom, so the micro-batch size was reduced to 1.

We trained in MXFP8, a Microscaling (MX) FP8 format~\citep{rouhani2023microscaling} in which each small block of consecutive elements (32 values) shares a single scaling factor, rather than applying one scale across an entire tensor as in conventional FP8~\citep{micikevicius2022fp8formatsdeeplearning}. This finer-grained, block-wise scaling better accommodates outliers in activations and gradients while retaining the throughput and memory benefits of 8-bit storage. We used the E4M3 format for both the forward and backward passes. To preserve numerical stability, the main model parameters and gradients were maintained in FP32, while the exponential moving averages and squared gradients used by the optimizer were stored in BF16. All remaining eligible tensors were represented in MXFP8. We further gather the distributed-optimizer parameter shards directly in FP8 and overlap this all-gather with computation: transmitting parameters in FP8 reduces the inter-GPU communication volume and ensures that the same low-precision weights are used in the gather and in the subsequent matrix multiplications---which aids the numerical stability of FP8 training---while overlapping the gather with compute hides its latency and improves throughput. The outlier-suppressed activation distribution induced by GN (Sec.~\ref{sec:architecture}) further supports stable low-precision training and is favorable for FP8 serving.

While training uses the MX-style block size of 32 elements, the released checkpoint is distributed in a \emph{blockwise} FP8 (E4M3) format with $128\times128$ weight blocks and dynamic per-token activation scaling, matching the recipe supported by mainstream FP8 serving kernels~\citep{deepgemm2025}. The model therefore serves in FP8 out of the box, with no separate post-hoc quantization step, and the same blockwise recipe is reused for RL rollouts (Sec.~\ref{sec:posttrain}).

\section{Post-Training}
\label{sec:posttrain}
The post-training pipeline consists of two SFT stages followed by multi-stage Reinforcement Learning (RL).
The SFT phase comprises an indexer SFT stage and a main SFT stage. Before the main post-training stages, we conduct a short indexer SFT, in which both the model parameters and the sparse-attention indexer parameters are trained. In subsequent stages, the indexer parameters are kept fixed and only the model parameters are updated.

\subsection{Supervised Fine-Tuning}\label{sec:data-sft}

Like its predecessor~\citep{skt2026axk1tr}, A.X K2 supports user-controllable hybrid inference, in which the thinking and non-thinking modes are explicitly specified at inference time.
When thinking mode is enabled, the response follows the format \texttt{<think>...</think>\{response\}}, following \citep{deepseekai2025deepseekr1, yang2025qwen3technicalreport}. When non-thinking mode is used, the response follows the format \texttt{</think>\{response\}}.

We observe that jointly training on thinking and non-thinking data can cause confusion between the two modes. In particular, thinking data, which is characterized by longer sample lengths, can dominate training and lead the model to generate thinking-mode responses even when non-thinking mode is requested.
Prior work has mitigated this issue by adopting a dual-track training strategy, as in our previous work~\citep{skt2026axk1tr}, or by adjusting the token ratio of reasoning to non-reasoning data to 1.5:1~\citep{bae2025exaone4}.
In this work, we instead unify training into a single track---our \textit{Think-Fusion} SFT recipe---and achieve reliable mode switching through data engineering. We construct a paired SFT dataset in which each prompt is associated with both a thinking and a non-thinking response. This prevents the model from relying on superficial distributional differences between thinking and non-thinking data and encourages it to follow the explicit mode instruction. We train A.X K2 on the paired SFT dataset, in which the thinking-to-non-thinking token ratio is approximately 13:1, and track mode confusion throughout post-training to verify that explicit control tokens remain effective after RL.

\subsubsection{Data}
The SFT dataset is organized into paired reasoning (thinking) and instruction (non-thinking) examples. The reasoning split emphasizes tasks that benefit from explicit intermediate reasoning, including mathematics, knowledge-intensive science questions, code, and agentic tool-use trajectories. The instruction split emphasizes concise instruction following, general chat, safety, and non-thinking counterparts for selected reasoning prompts. This paired construction lets the model learn mode control from the control tokens rather than from category-specific artifacts.

For safety data, A.X K1 already benefited from the safety SFT corpus introduced by ESSA~\citep{park2026essa}, which frames safety alignment as specification-guided deliberation over domain--task-conditioned safety requirements. In A.X K2, we further extend this source rather than simply reusing it: we run additional rounds of safety-specification evolution, label a larger English--Korean prompt pool with domain and task metadata, and synthesize new safety reasoning examples conditioned on the selected safety specification. This yields safety data that is better matched to the A.X K2 chat template and post-training mixture, while preserving the core ESSA objective of reducing both unsafe compliance and unnecessary over-refusal.

Concretely, each safety prompt is first mapped to one of 50 domain--task combinations, covering five application domains and ten response-task types. The resulting label is used only to select the corresponding evolved safety specification; the synthesis prompt itself exposes the selected specification and the user request, but not the internal domain--task label. The generated sample is then converted into the same chat-sample format used by the rest of the SFT pipeline, with the reasoning portion serialized for thinking-mode supervision and the final answer written as a natural user-facing response. We apply rubric-based filtering and targeted meta-review to remove samples with unsafe residue, language mismatch, formatting leakage, or excessive refusal, and to prefer safe completions that remain useful when a benign objective can be preserved.

Table~\ref{tab:sft-sample-ratio} shows the number of samples and ratio of the SFT dataset per category. The number of tokens and the distribution of token length are also provided in Appendix~\ref{appendix:sft_data}. 

\setlength{\tabcolsep}{5pt}
\begin{table}[htbp]
\centering
\small
\begin{tabular}{ll rr rr rr}
\toprule
 & & \multicolumn{2}{c}{Reasoning} & \multicolumn{2}{c}{Instruction} & \multicolumn{2}{c}{Total} \\
\cmidrule(lr){3-4}\cmidrule(lr){5-6}\cmidrule(lr){7-8}
Stage & Category & Samples & Ratio (\%) & Samples & Ratio (\%) & Samples & Ratio (\%) \\
\midrule

\multirow{8}{*}{Indexer SFT}
 & Math                  & 46,514    & 0.51  & 2,884     & 0.03  & 49,398    & 0.55 \\
 & Knowledge / Science   & 89,086    & 0.98  & 4,747     & 0.05  & 93,833    & 1.04 \\
 & Instruction Following & 5,275     & 0.06  & 621       & 0.01  & 5,896     & 0.07 \\
 & Code                  & 39,663    & 0.44  & 0         & 0.00  & 39,663    & 0.44 \\
 & Agent                 & 105,415   & 1.16  & 2,690     & 0.03  & 108,105   & 1.19 \\
 & General               & 73,061    & 0.81  & 64,317    & 0.71  & 137,378   & 1.52 \\
 & Safety                & 0         & 0.00  & 3,613     & 0.04  & 3,613     & 0.04 \\
\cmidrule(l){2-8}
 & \textbf{Sum}          & 359,014   & 3.97  & 78,872    & 0.87  & 437,886   & 4.84 \\
\midrule
\multirow{8}{*}{Main SFT}
 & Math                  & 585,678   & 6.47  & 90,247    & 1.00  & 675,925   & 7.47 \\
 & Knowledge / Science   & 1,355,439 & 14.98 & 148,375   & 1.64  & 1,503,814 & 16.62 \\
 & Instruction Following & 500,510   & 5.53  & 8,122     & 0.09  & 508,632   & 5.62 \\
 & Code                  & 496,978   & 5.49  & 65,144    & 0.72  & 562,122   & 6.21 \\
 & Agent                 & 1,602,902 & 17.71 & 35,993    & 0.40  & 1,638,895 & 18.11 \\
 & General               & 1,750,642 & 19.34 & 1,892,501 & 20.91 & 3,643,143 & 40.26 \\
 & Safety                & 34,083    & 0.38  & 45,164    & 0.50  & 79,247    & 0.88 \\
\cmidrule(l){2-8}
 & \textbf{Sum}          & 6,326,232 & 69.91 & 2,285,546 & 25.26 & 8,611,778 & 95.16 \\
\midrule
\multicolumn{2}{l}{\textbf{Total}}
 & 6,685,246 & 73.87 & 2,364,418 & 26.13 & 9,049,664 & 100.00 \\
\bottomrule
\end{tabular}
\caption{\textbf{Data composition by sample count and ratio (\%) across the SFT stages.}}
\label{tab:sft-sample-ratio}
\end{table}

\subsubsection{Training}

The SFT phase consists of two stages. First, we 
sample 0.4M examples without replacement from the SFT dataset and train the model together with the indexer so that the indexer learns the special control tokens. We then fix the indexer parameters and conduct the main SFT stage. For both stages, we train on packed 128K-token sequences, matching the model context length. For efficient data packing, we use a length-aware First-Fit Decreasing (FFD) bin-packing algorithm.
We sort all samples in a buffer in descending order of length and place each sample into the first bin that can accommodate it, opening a new bin only when none is available. By placing the largest samples first and filling the remaining gaps with smaller ones, this approach reduces unused space within each bin. As a result, the packing becomes more compact, yielding a higher effective-token ratio and thereby reducing padding waste. 

For both SFT stages, we minimize the standard negative log-likelihood loss. We use an indexer KL-loss coefficient of 0.1 during indexer SFT and set it to 0 thereafter.
For indexer training, we use a global batch size of 512 and a cosine learning-rate schedule with a peak of $1.1\times10^{-5}$ and a minimum of $2.2\times10^{-6}$, warming up for approximately 5\% of the total iterations. For main SFT, we reduce the global batch size to 128 and use a learning rate with a peak of $1.5\times10^{-5}$ and a minimum of $3.0\times10^{-6}$, with 2\% warmup iterations. As in pre-training, we use the global auxiliary loss with expert bias for expert load balancing and keep the hyperparameters unchanged. During training, we periodically evaluate performance and supplement the data accordingly. Table~\ref{tab:sft-sample-ratio} presents the full sample counts used for the SFT stages.

\begin{figure*}[t]
    \centering

    \begin{subfigure}[t]{0.50\textwidth}
        \centering
        \includegraphics[width=\linewidth]{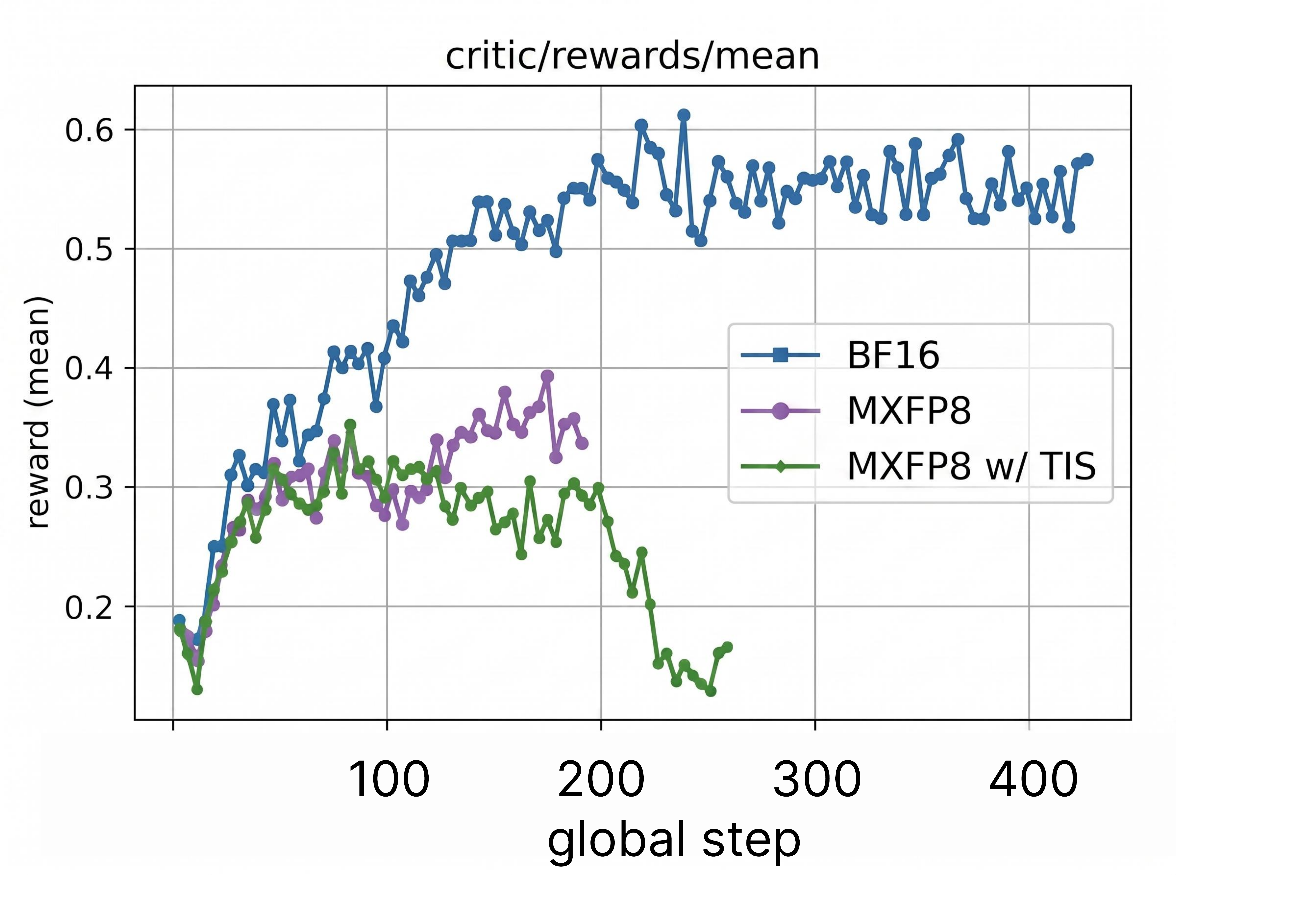}
        \caption{Training collapse under trainer--rollout precision mismatch.}
        \label{fig:fp8-stability-collapse}
    \end{subfigure}
    \hfill
    \begin{subfigure}[t]{0.46\textwidth}
        \centering
    \includegraphics[width=\linewidth]{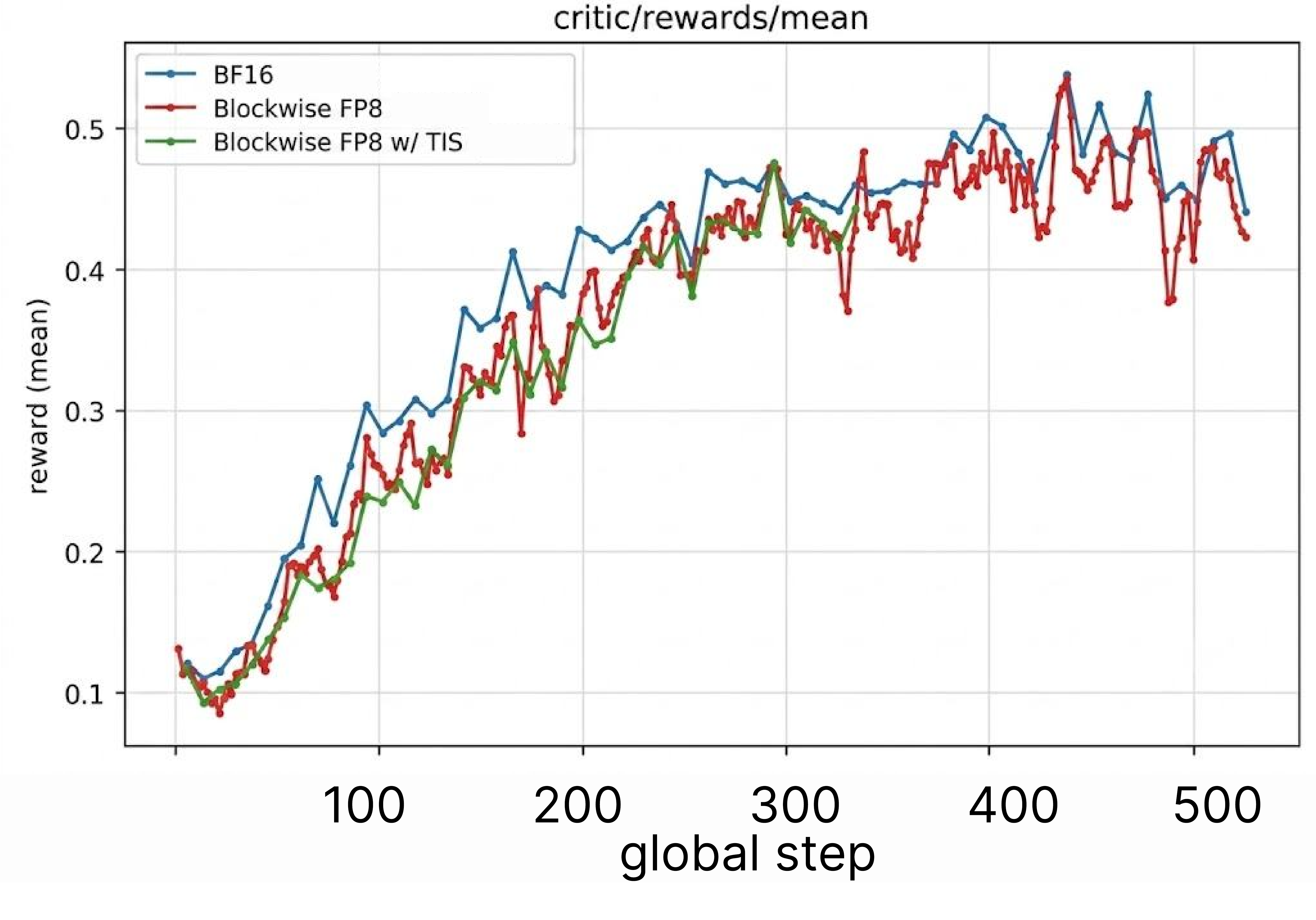}
        \caption{Stable training when trainer and rollout precision are matched.}
        \label{fig:fp8-stability-stable}
    \end{subfigure}

    \caption{
    \textbf{FP8 training stability depends on the trainer--rollout quantization consistency.} In all runs, the rollout engine, implemented with vLLM, uses the same blockwise FP8 format; only the trainer precision is varied. \textbf{Left:} When the trainer's precision is inconsistent with the rollout-side blockwise FP8 execution, reward learning becomes unstable and eventually collapses. Applying TIS does not prevent this collapse, indicating that token-level intervention alone cannot remove the underlying trainer--rollout precision mismatch. \textbf{Right:} When the trainer precision is aligned with the rollout-side blockwise FP8 format, training remains stable and avoids the collapse observed in the mismatched setting. These results suggest that stable low-precision RL requires matching the trainer's precision to the effective rollout precision.
    }
    \label{fig:fp8-stability}
\end{figure*}

\subsection{Reinforcement Learning Infrastructure}

Scaling RL to a model of A.X K2's size exposes two systems-level requirements that the architecture itself does not address: the training and rollout engines must agree numerically under low precision, and the rollout stage, which dominates wall-clock cost for a large Mixture-of-Experts (MoE) model, must be parallelized to match. We address these with an end-to-end blockwise-FP8 recipe and a data-parallel, expert-parallel asynchronous rollout pipeline, described below.

\paragraph{Blockwise FP8 for Trainer--Rollout Consistency.}
RL requires the training backend (Transformer Engine) and the rollout engine (vLLM) to operate under a numerically consistent FP8 recipe: when the two quantization schemes diverge, the sampling and update distributions drift apart, inflating the policy KL and destabilizing learning. While Blackwell (B200/B300) natively supports the microscaling \textsc{MXFP8} format, the vLLM rollout backend does not provide first-class MXFP8 support; its mature FP8 path for MoE modules instead targets the \emph{blockwise} FP8 recipe developed for Hopper, with per-block scaling and FP32 scale factors, as used by DeepGEMM~\citep{deepgemm2025}. To keep the trainer and rollout on a single recipe, we standardize on blockwise FP8 end-to-end. Because the Blackwell Transformer Engine defaults to MX-style scaling, we enable blockwise FP8 on Blackwell through a patched TE branch that forces FP32 block scales. Figure~\ref{fig:fp8-stability} contrasts the two configurations: a trainer running MXFP8 against a blockwise-FP8 rollout exhibits reward collapse, whereas the consistent blockwise-FP8 end-to-end recipe trains stably to convergence.

\begin{table}[t!]
    \centering
    \begin{tabular}{c|c}
    \toprule
    Parallel Configuration & Relative Throughput (Output tokens per second) \\
    \midrule
    TP8, DP1     & 1.0$\times$  \\
    TP1, DP8     & \textbf{1.6}$\times$  \\
    \bottomrule
    \end{tabular}
    \caption{\textbf{Relative throughput comparison between tensor-parallel and data-parallel configurations.} Throughput is normalized to the TP8, DP1 setting, showing that TP1, DP8 achieves 1.6$\times$ higher throughput under the same experimental setup (input length = 131{,}072 tokens).}
    \label{tab:throughput_comp}
\end{table}

\paragraph{Data-Parallel Asynchronous Rollout.}
Large-scale RL on an MoE model is dominated by rollout (generation) cost, so the rollout engine must be parallelized efficiently. The stock veRL pipeline~\citep{verl} is oriented toward small-model research and supports only single-replica rollout (DP1), which is adequate for dense models but leaves a large MoE such as A.X K2 severely throughput-bound: a single tensor-parallel replica serializes generation behind per-layer tensor-parallel all-reduces and underutilizes the cluster during the generation-heavy phase of each step. We instead develop a fully asynchronous rollout pipeline that shifts attention parallelism from tensor parallelism to data parallelism: rather than the trainer-style TP8/DP1 layout, the rollout engine runs TP1/DP8 while keeping expert parallelism fixed (EP8). This layout (i) maximizes MoE rollout throughput by replicating attention across eight data-parallel replicas and eliminating per-layer tensor-parallel all-reduces, (ii) keeps expert sharding identical to the trainer (EP8) so that policy weights can be resharded into the rollout engine cleanly, and (iii) decouples generation from optimization, allowing rollout and training to overlap. Together, these changes scale the rollout stage from the small-model regime to full-size MoE RL while preserving trainer--rollout consistency.

\subsection{Multi-Stage Reinforcement Learning}

We conduct RL over multiple stages, keeping the training setup fixed while varying the data composition across stages.
Rather than dedicating each stage to a distinct capability, every stage jointly optimizes instruction following, human-preference alignment, and agentic tool use under a shared reward framework; later stages additionally incorporate safety training.
We target these four capabilities because, unlike math and coding, they are particularly well suited to on-policy RL, which can directly optimize behaviors that emerge during model generation.  We also found that training a single capability with RL in isolation often led to over-optimization: the policy learned to exploit capability-specific reward signals while regressing on behaviors not represented in that stage.

\textbf{Data and reward design.} The training mixture is organized into four groups---instruction following, human preference, agentic tool use, and safety. Rather than fixing the mixture in advance, we treat it as a control surface across stages: using intermediate RL checkpoints, we repeatedly identified the model's weakest behaviors, synthesized targeted data to address them, and re-weighted or replaced datasets at each stage boundary. Thinking and non-thinking modes are sampled in roughly equal proportion throughout.

\begin{itemize}
    \item \textbf{Instruction following} uses rule-based verifiable rewards that check explicit constraints: format, length, required expressions, and schema conformance for structured-output prompts. We apply difficulty filtering to this group with an in-house small model, discarding prompts that the model already solves reliably so that the on-policy signal concentrates on informative cases.
    \item \textbf{Human preference} is scored by a pointwise LLM-as-judge: for each prompt we obtain a reference answer from a strong external model and supply the judge with few-shot pointwise scoring anchors. The judge first classifies the task into one of six domains (factual, reasoning, coding, extraction, creative, open) and then applies a domain-specific rubric over four axes (accuracy, completeness, clarity, helpfulness), with explicit safeguards against verbosity bias and reward hacking.
    \item \textbf{Agentic tool use} is rewarded in two complementary ways. Single-step tool-call data uses verifiable rewards that compare emitted calls against reference calls for schema conformance and argument correctness. Long-horizon agentic tasks use a gated judge: a binary gate first checks that the response constitutes a structurally valid, executable tool-calling transcript, assigning the minimum reward on failure, and only gate-passing responses are forwarded to the reference-based pointwise judge described above---preventing the judge from rewarding fluent responses that never issue a valid tool call.
    \item \textbf{Safety} data comprises red-teaming prompts, safety-preference data, and model-identity data. Red-teaming responses are scored by an LLM judge along nine safety dimensions---covering harm categories such as insults, adult content, illegal activity, hate, and bias, as well as informativeness and comprehension---with a holistic harmfulness rating determining the reward, so that the policy is rewarded for safe completions and penalized for harmful ones rather than merely for refusing. Safety-preference data is judged against principle-based rubrics, using the dataset's preferred and rejected responses as calibration references, and identity data is judged against rubrics encoding the model's identity policy.
\end{itemize}

In addition, for datasets prone to excessive verbosity, we apply the group-relative length penalty~\citep{he2025thinkdial}. Within each rollout group, shorter correct responses are rewarded and longer ones penalized, while incorrect responses are never rewarded for brevity.

\textbf{Algorithm and hyperparameters.} All stages share a single optimization recipe. We optimize with CISPO~\citep{minimax2025minimaxm1scalingtesttimecompute}, which clips the importance-sampling weight rather than the token-level update, so that low-probability but behavior-critical tokens continue to contribute gradient. Alongside the task reward, we use an auxiliary format reward and adopt GDPO~\citep{liu2026gdpo}, which normalizes each reward separately before combining them, preserving the resolution of each signal and improving multi-reward training stability. We do not apply a KL penalty and sample 16 rollouts per prompt at temperature 1.0, with a global batch size of 80 prompts and a learning rate of $2\times10^{-6}$ held constant after a brief warmup. In total, RL consumes roughly 100K prompts across all stages. Throughout, we retain the AdamW configuration used in pre-training but reduce the optimizer epsilon from $10^{-8}$ to $10^{-15}$, following the finding of MiniMax-M1~\citep{minimax2025minimaxm1scalingtesttimecompute} that a much smaller epsilon stabilizes optimization under the small gradient magnitudes characteristic of RL.

\section{Evaluation}
\label{sec:evaluation}

\subsection{Evaluation Settings}
We evaluate A.X K2---alongside its predecessor A.X K1 and leading open-weight baselines---on representative public benchmarks widely used in recent large language model evaluations, with particular attention to Korean-language tasks that are underrepresented in existing evaluations. Unless otherwise noted, all models are evaluated in thinking mode; A.X K2 additionally supports a non-thinking mode through its Think-Fusion recipe (Sec.~\ref{sec:posttrain}).

The evaluation is organized into six categories: Math, Korean, Code, Science \& Knowledge, General, and Agentic, each targeting a distinct aspect of model capability. Benchmarks marked with an asterisk (*) are evaluated by Artificial Analysis. Detailed descriptions of each benchmark and the evaluation method are provided in Appendices~\ref{subsec:details_of_evaluation_method} and \ref{subsec:details_of_evaluation_benchmarks}, and a complementary multilingual evaluation is reported in Appendix~\ref{appendix:globalmmlu}.

\subsection{Evaluation Results}

\begin{table}[t!]
\centering
\renewcommand{\arraystretch}{1.4}
\setlength{\tabcolsep}{4pt}
\begin{adjustbox}{width=1.0\textwidth,center}
\begin{tabular}{l|c|*{7}{c}}
\toprule
\textbf{Benchmark}
& \textbf{A.X K2}
& \textbf{A.X K1}
& \makecell{\textbf{Qwen3.5-} \\ \textbf{397B-A17B}}
& \makecell{\textbf{Nemotron 3} \\ \textbf{Ultra}}
& \makecell{\textbf{DeepSeek-V4} \\ \textbf{Flash}}
& \textbf{GLM-5.1}
& \textbf{Kimi-K2.6}
& \makecell{\textbf{MiniMax-} \\ \textbf{M2.7}} \\
\midrule

\# Total Params  & 688B & 519B & 397B & 550B & 284B & 754B & 1T  & 230B \\
\# Active Params & 33B  & 33B  & 17B  & 55B  & 13B  & 40B  & 32B & 10B  \\

\midrule
\multicolumn{9}{c}{\textit{\textbf{Math}}} \\
\addlinespace[4pt]
AIME26                    & \textbf{97.1} & 91.3 & 92.5 & 85.4 & 96.7 & 95.4 & 95.4 & 88.8 \\
Apex                      & \textbf{45.8} & 1.0  & 13.5 & 3.1  & 28.1 & 10.4 & 20.8 & 1.0  \\
Apex-shortlist            & \textbf{88.6} & 50.8 & 61.2 & 45.5 & 88.0 & 72.1 & 72.6 & 30.3 \\
KMO26 {\tiny(1st round)}  & 92.5 & 85.0 & 78.8 & 85.0 & \textbf{94.4} & 78.1 & 88.1 & 80.0 \\

\midrule
\multicolumn{9}{c}{\textit{\textbf{Korean}}} \\
\addlinespace[4pt]
KMMLU-Pro & \textbf{80.5} & 68.9 & 78.4 & 69.2 & 78.8 & 76.5 & 73.4 & 66.6 \\
KoBALT    & 73.0 & 51.0 & 69.9 & 59.1 & \textbf{75.3} & 72.0 & 66.0 & 51.4 \\
CLIcK     & \textbf{91.6} & 85.3 & 88.4 & 84.1 & 89.6 & 88.8 & 80.9 & 76.1 \\

\midrule
\multicolumn{9}{c}{\textit{\textbf{Code}}} \\
\addlinespace[4pt]
LiveCodeBench v6 {\tiny(Feb--May)} & 84.0 & 74.9 & 82.6 & 76.4 & \textbf{89.4} & 86.4 & 86.5 & 69.8 \\
SciCode                            & 41.0 & 24.1 & 42.0 & 39.9 & 44.9 & 43.8 & \textbf{53.5} & 47.0 \\
Terminal Bench v2.1*               & 36.0 & --   & 51.3 & 53.9 & 61.8 & 61.8 & \textbf{65.9} & 55.4 \\

\midrule
\multicolumn{9}{c}{\textit{\textbf{Science \& Knowledge}}} \\
\addlinespace[4pt]
Humanity's Last Exam & 27.8 & 8.3  & 27.3 & 26.6 & 32.1 & 28.0 & \textbf{35.9} & 28.1 \\
GPQA Diamond         & 85.6 & 76.4 & 89.3 & 86.7 & 89.4 & 86.8 & \textbf{91.1} & 87.4 \\
AA-Omniscience*      & 39.6 & 15.5 & 24.3 & 38.3 & 26.0 & 39.7 & \textbf{42.3} & 39.3 \\

\midrule
\multicolumn{9}{c}{\textit{\textbf{General}}} \\
\addlinespace[4pt]
IFBench & 75.9 & 63.2 & 78.8 & \textbf{81.4} & 79.2 & 76.3 & 76.0 & 75.7 \\
AA-LCR  & 66.0 & 26.0 & 65.7 & 67.0 & 63.0 & 62.3 & \textbf{69.7} & 68.7 \\

\midrule
\multicolumn{9}{c}{\textit{\textbf{Agentic}}} \\
\addlinespace[4pt]
GDPval* {\tiny(Elo)}                 & 1031 & 500 & 962  & 1164 & 1189 & \textbf{1257} & 1190 & 1158 \\
\(\tau^2\)-Bench* {\tiny(Telecom)}   & \textbf{98.0} & 86.0 & 95.6 & 83.3 & 95.0 & 97.7 & 95.9 & 84.8 \\
\(\tau^3\)-Bench* {\tiny(Banking)}   & 13.0 & --  & 13.4 & 13.8 & \textbf{22.9} & 11.5 & 20.6 & 8.9  \\
BrowseComp {\tiny($\leq$10 searches)} & 9.3 & --  & 26.9 & 13.4 & 16.8 & \textbf{29.1} & 21.5 & 14.2 \\

\bottomrule
\end{tabular}
\end{adjustbox}
\caption{\textbf{Comparison with strong open-weight LLMs.} The best score in each row is shown in bold. For benchmarks marked with *, the scores of the other open-weight models are taken from Artificial Analysis. GDPval is reported as an Elo rating; all other benchmark scores are reported as percentages.}
\label{tab:LLM_large_latest_think}
\end{table}

To validate the effectiveness of A.X K2, we conducted a comprehensive evaluation across the six categories of Math, Korean, Code, Science \& Knowledge, General, and Agentic capabilities. We compared our model against its predecessor A.X K1 and strong open-weight models---Qwen3.5-397B-A17B~\citep{qwen2026qwen35397b} (Qwen3.5 hereafter), Nemotron 3 Ultra~\citep{nvidia2026nemotron3ultra}, DeepSeek-V4 Flash~\citep{deepseekai2026deepseekv4}, GLM-5.1~\citep{zai2026glm51}, Kimi-K2.6~\citep{kimiteam2026kimik26}, and MiniMax-M2.7~\citep{minimax2026m27}---as shown in Table~\ref{tab:LLM_large_latest_think}.

A.X K2 is strongest in Math and Korean. In the Math domain, it achieves the best scores among all compared models on AIME26, Apex, and Apex-shortlist. Beyond final-answer benchmarks, we also tested A.X K2's proof-writing capability on IMO 2025 and the KMO26 2nd round, using the iterative proof refinement method of DeepSeekMath-V2~\citep{deepseekai2025deepseekmathv2}. A.X K2 scored 35/42 on IMO 2025 with a perfect 7/7 on each of the first five problems---reaching the gold-medal threshold of 35---and produced correct proofs for all 8 problems of the KMO26 2nd round. In the Korean domain, A.X K2 leads on KMMLU-Pro and CLIcK, and remains competitive on KoBALT, confirming its strength in the target language context. These results also reflect substantial gains over A.X K1 across every category, most notably on Apex (1.0 $\rightarrow$ 45.8), Humanity's Last Exam (8.3 $\rightarrow$ 27.8), and AA-LCR (26.0 $\rightarrow$ 66.0).

In the Science \& Knowledge and General categories, A.X K2 performs on par with strong open-weight models: it ranks among the top models on AA-Omniscience and matches the field on Humanity's Last Exam and AA-LCR, while GPQA Diamond and IFBench show a modest gap to the best baselines. In the Agentic category, despite only limited agentic reinforcement learning during post-training, A.X K2 attains a moderate level of agentic capability: it achieves the best score among all compared models on \(\tau^2\)-Bench Telecom, its GDPval Elo rating of 1031 is a large improvement over A.X K1 (500), and its \(\tau^3\)-Bench score is comparable to Qwen3.5 and Nemotron 3 Ultra. The remaining gaps on Terminal Bench and BrowseComp relative to models trained with heavier agentic optimization indicate clear headroom for future agentic post-training. Beyond Korean and English, a multilingual evaluation on Global-MMLU-Lite (Appendix~\ref{appendix:globalmmlu}) shows that A.X K2 improves over A.X K1 in all four evaluated languages while sharply reducing the variation across them.

\subsection{Long-Context Evaluation}
\label{subsec:long_context_eval}

A.X K2 is trained natively to a 128K context and serves with SGA sparse attention enabled (Sec.~\ref{subsec:long_context}), so its long-context behavior raises two distinct questions: how far the usable context extends, and whether replacing dense attention with a fixed top-$k$ budget costs quality. We evaluate both, and close with the serving efficiency that the sparse component buys.

\paragraph{Retrieval probing beyond the supported context.} A.X K2 is released with a supported context of 128K native and 256K under YaRN (Table~\ref{tab:ax}). To probe how far positional generalization extends past that, we run needle-in-a-haystack (NIAH) retrieval under zero-shot YaRN scaling with scaling factors of 1, 2, and 4---i.e., at 128K, 256K, and 512K---and observe a \emph{perfect} retrieval score at all three lengths and every needle depth, without any length-specific fine-tuning; Figure~\ref{fig:niah_256k} shows the 256K case. We report the 512K result as a retrieval-probing result only: NIAH measures retrieval rather than long-context understanding, and we do not claim 512K as a supported serving configuration. Notably, the same perfect retrieval holds after NVFP4 quantization (Sec.~\ref{subsec:low_precision}), so 4-bit serving does not erode long-range retrieval.

\paragraph{RULER.} Retrieval alone does not establish long-context \emph{understanding}, so we further evaluate on RULER~\citep{hsieh2024ruler}, which spans retrieval, multi-hop tracing, aggregation, and question-answering tasks across sequence lengths. As summarized in Table~\ref{tab:ruler}, A.X K2 reaches an overall average of 94.6 across lengths up to 256K, retaining 95.1\% of its 4K score at 128K and 88.8\% at 256K, twice the natively trained length. Because these scores are measured on the released sparse-attention model, they are obtained while each query reads only 2,048 positions, i.e., 1.6\% of the KV cache at 128K and 0.8\% at 256K.

\begin{table}[t!]
\centering
\small
\begin{tabular}{l *{7}{c} c}
\toprule
\textbf{Model} & \textbf{4K} & \textbf{8K} & \textbf{16K} & \textbf{32K} & \textbf{64K} & \textbf{128K} & \textbf{256K} & \textbf{Overall} \\
\midrule
A.X K2 & 97.5 & 97.2 & 97.5 & 96.5 & 94.3 & 92.7 & 86.6 & 94.6 \\
\bottomrule
\end{tabular}
\caption{\textbf{Long-context performance on RULER across sequence lengths.} Scores are measured on the released model with SGA sparse attention enabled. Lengths up to 128K are evaluated within the natively trained context; the 256K column applies YaRN scaling with a scaling factor of 2 (Sec.~\ref{subsec:long_context}).}
\label{tab:ruler}
\end{table}

\paragraph{LongBench v2.} RULER is synthetic by construction, so we complement it with LongBench v2~\citep{bai2025longbenchv2}, a four-way multiple-choice benchmark with inputs of up to 200K tokens drawn from documents, academic papers, code repositories, structured data, and dialogue histories. Although its authors state that all test data are in English, 87 of the 503 items (17.3\%) contain Han characters (CJK ideographs). These occur in Chinese or Japanese passages, code comments, and glossary entries in the low-resource translation task. We remove these items from every model's evaluation, leaving 416. Table~\ref{tab:longbench-v2} shows that A.X K2 scores 63.9 overall, compared with 64.2 for Qwen3.5 and 63.2 for GLM-5.1. The three models tie at 56.2 on the \textbf{Long} split, while A.X K2 has the highest score on the \textbf{Easy} split (74.7). Because the filter changes the model ranking, we also report results on all 503 items in Table~\ref{tab:longbench-v2-full} of Appendix~\ref{appendix:longbenchv2-full}. On the full set, A.X K2 scores 62.2 and ranks third, behind Qwen3.5 (64.6) and GLM-5.1 (63.8). Its score increases by 1.7 points after filtering, whereas the scores of the other three models decrease by 0.1--0.6 points. As with RULER, these results are obtained with the released sparse-attention model under the same sub-2\% attention budget.

\begin{table}[t!]
\centering
\small
\begin{tabular}{l c cc ccc}
\toprule
\textbf{Model} & \textbf{Overall} & \textbf{Easy} & \textbf{Hard} & \textbf{Short} & \textbf{Medium} & \textbf{Long} \\
\midrule
A.X K2           & 63.9          & \textbf{74.7} & 57.9          & 69.2          & 62.8          & \textbf{56.2} \\
Qwen3.5-397B-A17B & \textbf{64.2} & 72.7         & \textbf{59.4} & 67.3          & \textbf{65.0} & \textbf{56.2} \\
GLM-5.1          & 63.2          & 70.0          & \textbf{59.4} & \textbf{70.5} & 60.0          & \textbf{56.2} \\
Nemotron 3 Ultra & 61.3          & 69.3          & 56.8          & 65.4          & 61.1          & 53.8          \\
\bottomrule
\end{tabular}
\caption{\textbf{LongBench v2 accuracy (\%) on the 416-item filtered set.} We remove 87 of the 503 items (17.3\%) because they contain Han characters. The remaining set splits into 150 Easy and 266 Hard items by difficulty, and into 156 Short, 180 Medium, and 80 Long items by input length. The filter is applied identically to every model, but its effect varies by task category: it removes 40\% of the Code Repository Understanding items and none of the Long-dialogue History Understanding items. We use the official LongBench v2 harness with its zero-shot template, middle truncation to each model's context limit, and unmodified scorer; thinking is enabled. A.X K2 is evaluated with SGA sparse attention and FP8 weights, using a 200K input budget under YaRN scaling with a scaling factor of 2 (Sec.~\ref{subsec:long_context}); the baselines are served with the setup described in Appendix~\ref{subsec:details_of_evaluation_method}. The highest model score in each column is shown in bold, including ties.}
\label{tab:longbench-v2}
\end{table}

\paragraph{Sparsity is quality-neutral.} To isolate the effect of the sparse component itself, we compare the model immediately before the SGA adaptation stage---which still runs dense attention over the full 128K context---with the released model after adaptation, on LongBench v1~\citep{bai2024longbench}. As shown in Table~\ref{tab:sga-longbench}, the score does not drop: 62.80 dense versus 62.99 sparse ($+0.19$). Pruning attention to 1.6\% of the context at 128K therefore leaves long-context quality unchanged within measurement noise, so the efficiency gains reported below are obtained without a measured quality cost.

\begin{table}[t!]
\centering
\small
\begin{tabular}{l l c}
\toprule
\textbf{Stage} & \textbf{Attention} & \textbf{LongBench v1} \\
\midrule
Before SGA adaptation (Stage~3B) & Dense (full)                 & 62.80 \\
After SGA adaptation (Stage~3C)  & Sparse (top-$k$, $k{=}2048$) & 62.99 \\
\midrule
\multicolumn{2}{l}{$\Delta$ (sparse $-$ dense)} & $+0.19$ \\
\bottomrule
\end{tabular}
\caption{\textbf{Effect of the SGA sparse component on long-context quality.} Both checkpoints support a 128K context; they differ only in whether attention is dense or restricted to the indexer's top-$k$ selection. Introducing sparsity leaves LongBench v1 unchanged within noise.}
\label{tab:sga-longbench}
\end{table}

\paragraph{Long-context reasoning and Korean long-context tasks.} Two further results corroborate this picture. On AA-LCR (Table~\ref{tab:LLM_large_latest_think}), which tests coherence and information retention over long inputs, A.X K2 scores 66.0, ahead of DeepSeek-V4 Flash (63.0) and GLM-5.1 (62.3), and 40 points above A.X K1 (26.0). On the Long Context category of our in-house KS-Eval (Sec.~\ref{sec:ks-eval}), which evaluates Korean long-context tasks over 8 tasks and 490 items, A.X K2 improves by 27.11~pp over A.X K1---a 68.2\% relative gain---and leads Qwen3.5 by 3.70~pp, while GLM-5.1 leads the category.

\paragraph{Efficiency.} Because the top-$k$ budget is fixed while the context grows, the benefit of SGA widens with input length. In long-context serving, A.X K2 sustains substantially higher total-token throughput than A.X K1 as the input scales toward 120K tokens (Figure~\ref{fig:throughput-k1-k2}; analyzed in Sec.~\ref{subsec:serving_throughput}), and the reduced KV-cache footprint of MLA combines with FP8/NVFP4 weights to lower the memory required to serve those contexts (Sec.~\ref{subsec:low_precision}). Taken together with the quality results above, A.X K2 reaches 94.6 overall on RULER out to 256K while reading under 2\% of the context per query.

\subsection{Korean-Centric Evaluations}

\subsubsection{KS-Eval}
\label{sec:ks-eval}

\paragraph{Motivation and design.}
Public benchmarks are useful for comparing general capabilities, but most are
designed around English-language tasks and may not capture performance
differences arising from Korean linguistic conventions, culturally grounded
knowledge, or the interaction patterns of Korean-language service
environments. KS-Eval is an in-house benchmark that complements public
evaluations by targeting capabilities frequently exercised in practical
services. It spans four categories: Common Sense (knowledge and reasoning
over everyday situations), Instruction Following (adherence to user
requirements, multi-step instructions, and output constraints), Long Context
(information retention and reasoning over extended inputs), and STEM
(mathematical, scientific, and technical problem solving). Tasks incorporate
Korean linguistic characteristics and cultural context, and the
multidimensional structure gives a more detailed view of model strengths than
a single aggregate score.

\paragraph{Evaluation scope and protocol.}
The suite used in this report comprises 26 tasks and 1,800 items (Common
Sense: 4 tasks / 496 items; Instruction Following: 8 / 426; Long Context:
8 / 490; STEM: 6 / 388). We compare A.X K2 against its predecessor A.X K1 and
two strong external baselines, Qwen3.5 and GLM-5.1, all evaluated in thinking
mode using the same pipeline and task data. This controlled setting is
intended to measure generation-over-generation improvement and assess
competitiveness rather than to provide an exhaustive leaderboard. For each
task we use the first available metric among \texttt{accuracy},
\texttt{f1\_score}, \texttt{rouge\_l\_f1}, \texttt{item\_pass\_rate}, and
\texttt{score}, normalized to a 0--100 scale. Each category score is the
unweighted macro average over its constituent tasks, so item counts describe
the scale of the evaluation but do not affect aggregation.

\begin{table}[t]
    \centering
    \small
    \setlength{\tabcolsep}{3.5pt}
    \begin{tabular*}{\linewidth}{@{\extracolsep{\fill}}lrrrrrr@{}}
        \toprule
        \textbf{Category} & \textbf{\# Tasks} & \textbf{\# Items} &
        \textbf{A.X K2} &
        \textbf{A.X K1} &
        \makecell{\textbf{Qwen3.5-}\\\textbf{397B-A17B}} &
        \textbf{GLM-5.1} \\
        \midrule
        Common Sense
        & 4 & 496 & \textbf{57.45} & 37.98 & 56.40 & 47.89 \\
        Instruction Following
        & 8 & 426 & \textbf{71.67} & 43.69 & 67.88 & 71.07 \\
        Long Context
        & 8 & 490 & 66.85 & 39.74 & 63.15 & \textbf{72.49} \\
        STEM
        & 6 & 388 & 75.47 & 62.01 & \textbf{83.10} & 79.77 \\
        \midrule
        Total
        & 26 & 1,800 & -- & -- & -- & -- \\
        \bottomrule
    \end{tabular*}
    \caption{\textbf{Comparison on KS-Eval across four capability categories
    (thinking mode).} Scores are percentages; each category score is the
    unweighted macro average over its constituent tasks and is not weighted
    by item count. The best score in each row is shown in bold. Dashes in the
    \textit{Total} row indicate that no single cross-category aggregate is
    reported; the row summarizes evaluation scope only.}
    \label{tab:kseval}
\end{table}

\paragraph{Results.}
As shown in Table~\ref{tab:kseval}, A.X K2 improves over A.X K1 in all four
categories: 19.47 percentage points (pp) in Common Sense, 27.98~pp in
Instruction Following, 27.11~pp in Long Context, and 13.46~pp in STEM. The
Instruction Following and Long Context gains correspond to relative
improvements of approximately 64.0\% and 68.2\%. A.X K2 also scores higher
than A.X K1 on all 26 individual tasks, indicating that the improvement is
broadly distributed rather than driven by a small subset of tasks.

Among the compared models, A.X K2 ranks first in Common Sense (57.45\%;
+1.05~pp over Qwen3.5, +9.56~pp over GLM-5.1) and Instruction Following
(71.67\%; +3.79~pp over Qwen3.5 and +0.60~pp over GLM-5.1). The Instruction
Following result is particularly relevant to practical deployment, where
reliable adherence to complex user instructions and output constraints is a
foundational capability. A.X K2 remains competitive in the remaining categories,
outperforming Qwen3.5 in Long Context (66.85\% vs.\ 63.15\%) while GLM-5.1
leads at 72.49\%, and reaching 75.47\% in STEM against 79.77\% (GLM-5.1) and
83.10\% (Qwen3.5).

\paragraph{Evaluation--development cycle.}
KS-Eval serves not only as a measurement tool but also as part of an iterative
cycle in which evaluation results directly inform later model improvements.
After each evaluation, task-level success and failure patterns are analyzed to
identify model strengths and the capabilities that require further refinement.
These findings are incorporated into training data construction and subsequent
model training, and the updated model is re-evaluated on KS-Eval. The process
targets both Korean-specific linguistic and cultural understanding and the
capabilities required in real-world service scenarios, including instruction
following, long-context processing, knowledge utilization, and problem solving.
The consistent gains of A.X K2 over A.X K1 reported above are the product of
this cycle of evaluation, analysis, refinement, and re-evaluation.

\subsubsection{Manufacturing Benchmark}
\label{sec:manufacturing-bench}

\paragraph{Motivation and design.}
We evaluate A.X K2 on the Manufacturing Benchmark, an in-house benchmark of Korean manufacturing knowledge built to assess readiness for deployment in a sector central to the Korean economy. It spans semiconductors, automotive, and chemicals, and measures two capabilities: command of the domain, and abstention when a question cannot be answered from the information given---together with whether both hold up once a misleading cue is introduced. Questions are written at the level of specialized technical textbooks and grounded in Korean Industrial Standards (KS) documents, and each is presented in a closed-ended format with six substantive answer
choices plus a seventh abstention option. Of the 836 items, 670 are answerable, covering concept understanding, calculation, comparison, and compliance with regulations and standards. The remaining 166 are unanswerable---they carry a false premise, lack information needed for a unique answer, or are ambiguous in intent---and the intended behavior is to withhold rather than guess.

\paragraph{Evaluation scope and protocol.}
We evaluate three conditions. The baseline presents each question with no misleading cue. The \emph{user-assertion} condition introduces a user who claims that a particular incorrect option is correct, measuring whether the model maintains its judgment instead of deferring. The \emph{document-assertion} condition supplies a reference document that misstates the answer it supports, measuring whether the model examines the evidence rather than deferring to the source. The baselines are GLM-5.1, Qwen3.5, Qwen3.6-27B~\citep{qwen2026qwen3627b}, and both DeepSeek-V4 variants, Pro and Flash~\citep{deepseekai2026deepseekv4}.
Manufacturing deployments often operate under practical serving constraints
that favor smaller models; we therefore include Qwen3.6-27B in this benchmark
only, as a compact reference point. All models are evaluated in thinking mode.

Answerable items outnumber unanswerable ones, so accuracy alone would favor a model that answers everything. We therefore score the two behaviors separately: answerable F1 takes a correct answer as the target class, abstention F1 takes a correct withholding as the target class, and overall F1 is their arithmetic mean, weighting the two abilities equally. Needless abstentions are penalized in both. The same overall F1 is computed for each misleading-cue condition.

\begin{table}[t]
\centering
\small
\sisetup{detect-weight=true, detect-inline-weight=math}
\begin{tabular}{l
    S[table-format=2.1] S[table-format=2.1] S[table-format=2.1]
    S[table-format=2.1] S[table-format=2.1]}
\toprule
 & \multicolumn{3}{c}{Baseline} & \multicolumn{2}{c}{Misleading cue} \\
\cmidrule(lr){2-4} \cmidrule(lr){5-6}
Model & {Answerable} & {Abstention} & {Overall} & {User} & {Document} \\
\midrule
A.X K2               & 76.9           & \bfseries 61.1 & \bfseries 69.0 & \bfseries 67.8 & \bfseries 41.4 \\
Qwen3.5-397B-A17B    & \bfseries 80.6 & 54.2           & 67.4           & 56.9           & 7.8 \\
GLM-5.1              & 77.4           & 51.7           & 64.5           & 60.1           & 33.4 \\
DeepSeek-V4 Pro      & 78.6           & 49.1           & 63.8           & 60.9           & 20.4 \\
Qwen3.6-27B          & 76.4           & 37.6           & 57.0           & 54.5           & 18.5 \\
DeepSeek-V4 Flash    & 74.0           & 36.1           & 55.0           & 52.4           & 39.7 \\
\bottomrule
\end{tabular}
\caption{\textbf{Manufacturing Benchmark results.}
All scores are F1 (\%), so higher values are better; rows are ordered by
baseline overall F1. Every condition is evaluated over all 836 items:
answerable F1 is computed over the 670 answerable items, abstention F1 over
the 166 unanswerable items, and overall F1 is their arithmetic mean. The three
leftmost score columns report these quantities without a misleading cue; the
two rightmost report overall F1 recomputed when a false claim is introduced by
the user or by the document, so higher values indicate greater robustness to a
misleading cue. The highest value in each column is shown in bold.}
\label{tab:manufacturing-main}
\end{table}

\paragraph{Results.}
A.X~K2 records the highest overall F1 in Table~\ref{tab:manufacturing-main}, at 69.0\%, with Qwen3.5 nearest at 67.4\% and the remaining models 4.5 to 14.0 points behind. The lead does not come from answering: its answerable F1 of 76.9\% ranks fourth of six. It comes from abstention, where 61.1\% stands 6.9 points clear of the next model. The baselines convert unanswerable items into confident guesses, gaining a little on the answerable set and forfeiting far more on the rest.

A.X~K2's lead persists under adversarial pressure, although the ordering beneath it does not. When a user asserts an incorrect option, A.X~K2 retains 67.8\%, 6.9 points above the closest baseline (DeepSeek-V4 Pro) and 15.4 above the weakest (DeepSeek-V4 Flash). When a document is presented as supporting the wrong answer, it again leads, at 41.4\%, ahead of DeepSeek-V4 Flash (39.7\%) and GLM-5.1 (33.4\%). The baselines spread far more widely in this second condition---from 39.7\% down to 7.8\% for Qwen3.5---and they reorder: DeepSeek-V4 Flash places last under user assertion yet second under document assertion. Resistance to a user's insistence therefore predicts little about resistance to a fabricated citation.

At 41.4\%, the document-assertion score falls far below the 67.8\% A.X~K2 reaches under user assertion: a false claim is much harder to resist when it arrives in a supplied document than when the user makes it. Since industrial deployments lean heavily on supplied references, closing this gap is our priority for further work. Because the Manufacturing Benchmark scores domain expertise, abstention under uncertainty, and misleading-cue resistance separately, weak cases can be traced to a specific capability and fed back into training.

\subsubsection{Red-Teaming Benchmark}

\paragraph{Motivation and design.}
To assess safety under adversarial pressure, we evaluate A.X K2 on an in-house
red-teaming benchmark of 224 Korean adversarial prompts spanning five harm
domains (harmful content, malicious use, unfair representation, information and
safety violations, and misinformation and manipulation). The prompts were authored by our internal annotators and collected at an in-house red-teaming event, and 78\% of them are multi-turn conversations that build toward the harmful objective over several exchanges (median six messages, longest 78) rather than single-turn requests. We compare A.X K2 against its
predecessor A.X K1 and the strong contemporary baselines Qwen3.5 and
GLM-5.1, in both non-thinking and thinking inference modes.

\paragraph{Evaluation scope and protocol.}
As in KS-Eval, every model is evaluated directly under the same inference configuration, and each response is scored by the same automated judge---GPT-5 under a fixed rubric---which labels the response as a successful attack or not; we report the attack success rate (ASR), so lower is safer. Responses failing the rubric's instruction-following check count as unsuccessful attacks; the few requests that return no response count as successes. All rates are over the full 224 prompts.

\paragraph{Results.}
As shown in Table~\ref{tab:safety-by-domain}, A.X K2 is the most robust model
in non-thinking mode, with an ASR of 12.1\%---far below A.X K1 (59.4\%),
Qwen3.5 (21.9\%), and GLM-5.1 (66.1\%), and a substantial safety gain
over its predecessor. In thinking mode, A.X K2's ASR rises to 33.9\% while
Qwen3.5 attains the lowest ASR (21.0\%); A.X K2 nonetheless remains
well ahead of A.X K1 (65.2\%) and GLM-5.1 (40.6\%). The increase from
non-thinking to thinking mode indicates that extended reasoning can increase susceptibility to adversarial prompts, a trade-off we continue to address in post-training.

Notably, this mode effect is not shared across models. Moving from non-thinking
to thinking changes ASR by $+21.8$ points for A.X K2 and $+5.8$ for A.X K1, but
leaves Qwen3.5 essentially unchanged ($-0.9$) and \emph{reduces}
GLM-5.1 by $25.5$ points. Extended reasoning is therefore not inherently
safer or less safe; whether it helps depends on how safety behavior is
distributed between a model's reasoning trace and its final response.

\begin{table}[t]
    \centering
    \small
    \setlength{\tabcolsep}{4.5pt}
    \begin{tabular*}{\linewidth}
    {@{\extracolsep{\fill}}lrrrrr@{}}
        \toprule
        \textbf{Harm domain} & $\bm{n}$ & \textbf{A.X K2} & \textbf{A.X K1} & \makecell{\textbf{Qwen3.5-}\\\textbf{397B-A17B}} & \textbf{GLM-5.1} \\
        \midrule
        \multicolumn{6}{@{}l}{\emph{Non-thinking}} \\
        Harmful content            & 58 & \textbf{10.3} & 72.4 & 22.4 & 72.4 \\
        Malicious use              & 57 & \textbf{15.8} & 61.4 & 22.8 & 71.9 \\
        Unfair representation      & 30 & \textbf{16.7} & 56.7 & 33.3 & 80.0 \\
        Information/safety viol.   & 28 & \textbf{10.7} & 53.6 & 25.0 & 57.1 \\
        Misinformation/manip.      & 51 & \textbf{7.8}  & 47.1 & 11.8 & 49.0 \\
        \cmidrule(l{2pt}){1-6}
        Overall                    & 224 & \textbf{12.1} & 59.4 & 21.9 & 66.1 \\
        \midrule
        \multicolumn{6}{@{}l}{\emph{Thinking}} \\
        Harmful content            & 58 & 37.9 & 70.7 & \textbf{24.1} & 34.5 \\
        Malicious use              & 57 & 52.6 & 75.4 & \textbf{24.6} & 42.1 \\
        Unfair representation      & 30 & \textbf{26.7} & 63.3 & 33.3 & 56.7 \\
        Information/safety viol.   & 28 & \textbf{14.3} & 53.6 & 17.9 & 35.7 \\
        Misinformation/manip.      & 51 & 23.5 & 54.9 & \textbf{7.8} & 39.2 \\
        \cmidrule(l{2pt}){1-6}
        Overall                    & 224 & 33.9 & 65.2 & \textbf{21.0} & 40.6 \\
        \bottomrule
    \end{tabular*}
    \caption{\textbf{Red-teaming benchmark results by harm domain.}
    Scores are attack success rates (ASR; \%), where lower values
    indicate greater resistance to adversarial attacks. $n$ is the
    number of adversarial prompts in each domain. The lowest ASR in
    each row is shown in bold.}
    \label{tab:safety-by-domain}
\end{table}

\section{Inference Optimization}
\label{sec:inference_optimization}

The preceding sections established what A.X K2 can do; this section examines what it costs to serve. We first show that low-precision deployment preserves model quality (Sec.~\ref{subsec:low_precision}), and then quantify long-context serving throughput under a layered optimization strategy that applies decoding acceleration and quantization selectively on top of the SGA architecture (Sec.~\ref{subsec:serving_throughput}).

\subsection{Low-Precision Robustness}
\label{subsec:low_precision}

The outlier suppression provided by GN and the SGA output gate (Section~\ref{sec:architecture}) makes A.X K2 robust under low-precision deployment. Training in FP8 reduces the minimum serving-memory footprint to about 62\% of A.X K1's BF16 requirement, and NVFP4 quantization further reduces it to roughly 57\% of the FP8 model---about 36\% of A.X K1 (Table~\ref{tab:precision-memory}). The 276\,GB freed relative to FP8 is directly useful in long-context serving: at a fixed node count it can be redirected to KV cache, raising the concurrency sustainable at 128K--256K inputs.

\begin{table}[t]
\centering
\small
\begin{tabular}{l l r}
\toprule
\textbf{Model} & \textbf{Precision} & \textbf{Min.\ Memory} \\
\midrule
A.X K1        & BF16  & 1038\,GB \\
A.X K2        & FP8   & 646\,GB \\
A.X K2        & NVFP4 & 370\,GB \\
\bottomrule
\end{tabular}
\caption{\textbf{Minimum serving memory by model precision.}}
\label{tab:precision-memory}
\end{table}

Crucially, this compression incurs almost no quality loss. Table~\ref{tab:fp8-nvfp4} compares FP8 and NVFP4 across eleven representative Korean and English base-model tasks, using the same experts-only W4A4 configuration as in Figure~\ref{fig:niah_256k}. Averaged over the suite, NVFP4 scores 78.19 against 78.95 for FP8---a drop of 0.76 points, or 99.0\% of FP8 accuracy retained. The degradation is also well contained rather than concentrated in a single failure: nine of eleven tasks move by less than 2 points, three tasks (KoBEST-COPA, HumanEval, KLUE-MRC) \emph{improve} under NVFP4, and the largest regressions are on the two arithmetic-heavy tasks, GSM8K ($-2.50$) and MATH ($-2.42$). Long-context behavior is likewise preserved: NVFP4 retains a perfect NIAH score at 128K, 256K, and 512K (Sec.~\ref{subsec:long_context_eval}), so the 4-bit model gives up neither short-task accuracy nor long-range retrieval. This robustness is what the outlier suppression of GN and the SGA output gate is intended to deliver: narrow block-scaled formats such as NVFP4 share one scale across a small block, so a single massive activation would otherwise collapse the precision of every value co-located with it (Sec.~\ref{subsec:gated_blocks}). On dedicated NPU hardware, A.X K2 further attains a 107\% performance-per-watt ratio relative to a comparable GPU (NVIDIA L40S), measured on Rebellions ATOM-Max.

\begin{table}[t]
\centering
\small
\begin{adjustbox}{width=1.0\textwidth,center}
\begin{tabular}{l *{11}{c} c}
\toprule
\textbf{Precision} & \textbf{CLIcK} & \textbf{GSM8K} & \textbf{MMLU} & \makecell{\textbf{MMLU}\\\textbf{-Pro}} & \textbf{KMMLU} & \makecell{\textbf{KoBEST}\\\textbf{BoolQ}} & \makecell{\textbf{KoBEST}\\\textbf{COPA}} & \makecell{\textbf{MATH}} & \textbf{MBPP} & \makecell{\textbf{Human}\\\textbf{Eval}} & \makecell{\textbf{KLUE}\\\textbf{-MRC}} & \textbf{Avg.} \\
\midrule
FP8   & 84.21 & 80.21 & 82.27 & 69.71 & 78.41 & 96.72 & 88.30 & 60.50 & 72.00 & 79.88 & 76.20 & 78.95 \\
NVFP4 & 84.06 & 77.71 & 82.00 & 68.04 & 77.70 & 96.01 & 88.60 & 58.08 & 70.40 & 81.10 & 76.40 & 78.19 \\
\midrule
$\Delta$ & $-0.15$ & $-2.50$ & $-0.27$ & $-1.67$ & $-0.71$ & $-0.71$ & $+0.30$ & $-2.42$ & $-1.60$ & $+1.22$ & $+0.20$ & $-0.76$ \\
\bottomrule
\end{tabular}
\end{adjustbox}
\caption{\textbf{Base-model task accuracy under FP8 vs.\ NVFP4 quantization.} NVFP4 uses the experts-only W4A4 configuration. \textbf{Avg.} is the unweighted mean over the eleven tasks, and $\Delta$ is NVFP4 minus FP8, so negative values indicate a loss from 4-bit quantization. NVFP4 retains 99.0\% of the FP8 average.}
\label{tab:fp8-nvfp4}
\end{table}

\subsection{Long-Context Serving Throughput}
\label{subsec:serving_throughput}

Beyond quality retention, the practical question for a 688B model is whether it can be served more cheaply than its 519B predecessor. A.X K2 answers this through a \emph{layered} optimization strategy: on top of the SGA architecture (Sec.~\ref{sec:architecture}), decoding acceleration via EAGLE3 speculative decoding~\citep{li2025eagle3} and NVFP4 post-training quantization (Sec.~\ref{subsec:low_precision}) can each be applied selectively, depending on the workload. Measured against A.X K1 under identical conditions---a single B200 node, concurrency 32, output length fixed at 1K tokens, and input sequence length (ISL) swept from 1K to 120K---each layer moves the point at which A.X K2 overtakes A.X K1 in total-token throughput progressively earlier: 64K with SGA alone, 32K with EAGLE3 added, and \emph{every} input length from 1K upward under NVFP4 (Table~\ref{tab:throughput-vs-k1}). The advantage is largest precisely in the agentic long-context regime (32K+) that coding, tool use, and accumulated interaction histories occupy.

\begin{table}[t]
\centering
\small
\begin{tabular}{l r r r c}
\toprule
\textbf{Configuration} & \textbf{32K} & \textbf{64K} & \textbf{120K} & \textbf{Crossover vs.\ A.X K1} \\
\midrule
A.X K2 (FP8)      & $-18.2\%$ & $+2.7\%$  & $+35.4\%$  & 64K \\
\quad + EAGLE3    & $+4.4\%$  & $+33.0\%$ & $+65.9\%$  & 32K \\
NVFP4 PTQ         & $+37.8\%$ & $+69.4\%$ & $+123.7\%$ & all lengths (1K$\sim$) \\
\bottomrule
\end{tabular}
\caption{\textbf{Total-token throughput of A.X K2 serving configurations relative to A.X K1} (B200 single node, concurrency 32, OSL 1K). EAGLE3 is applied on top of the FP8 configuration, whereas NVFP4 PTQ is a separate configuration rather than a further layer on EAGLE3. \textbf{Crossover} is the input length from which the configuration exceeds A.X K1.}
\label{tab:throughput-vs-k1}
\end{table}

\begin{figure}[t!]
    \centering
    \includegraphics[width=\textwidth]{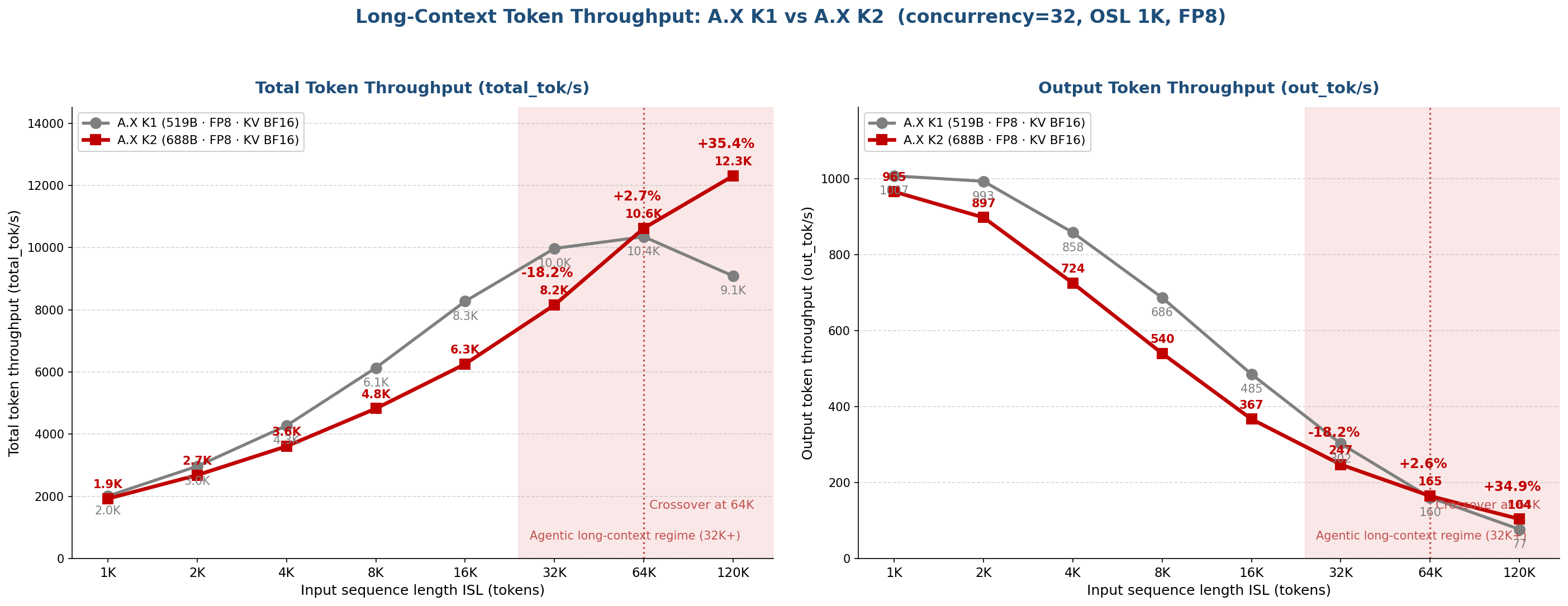}
    \caption{\textbf{Long-context serving throughput of A.X K2 vs.\ A.X K1 under matched FP8 precision} (B200 single node, concurrency 32, OSL 1K, ISL 1K--120K). \textbf{Left:} total-token throughput; \textbf{right:} output-token throughput. A.X K1 peaks at 64K and declines thereafter, while A.X K2 keeps rising through 120K, crossing over at 64K; the shaded band marks the agentic long-context regime (32K+).}
    \label{fig:throughput-k1-k2}
\end{figure}

\paragraph{Baseline: A.X K1 vs.\ A.X K2.} Under matched precision (FP8 weights, BF16 KV cache), A.X K2 overtakes A.X K1 from 64K inputs onward, and the gap widens as the input grows: $+2.7\%$ at 64K (10.6K total tok/s) and $+35.4\%$ at 120K (12.3K vs.\ 9.1K). The two curves in Figure~\ref{fig:throughput-k1-k2} diverge for a structural reason: A.X K1 peaks at 64K (10.4K tok/s) and then \emph{declines} to 9.1K at 120K, as the attention computation of dense attention grows steeply with input length---a regime in which throughput falls as inputs get longer. A.X K2's sparse attention instead keeps the attention cost per token nearly constant, so throughput continues to rise through 120K, securing long-context scalability at the architecture level. Below 32K, where attention is not yet the bottleneck, the larger parameter count (519B $\rightarrow$ 688B) leaves A.X K2 behind ($-18.2\%$ at 32K); this short-input deficit is what the decoding and quantization layers below recover.

\begin{figure}[t!]
    \centering
    \includegraphics[width=\textwidth]{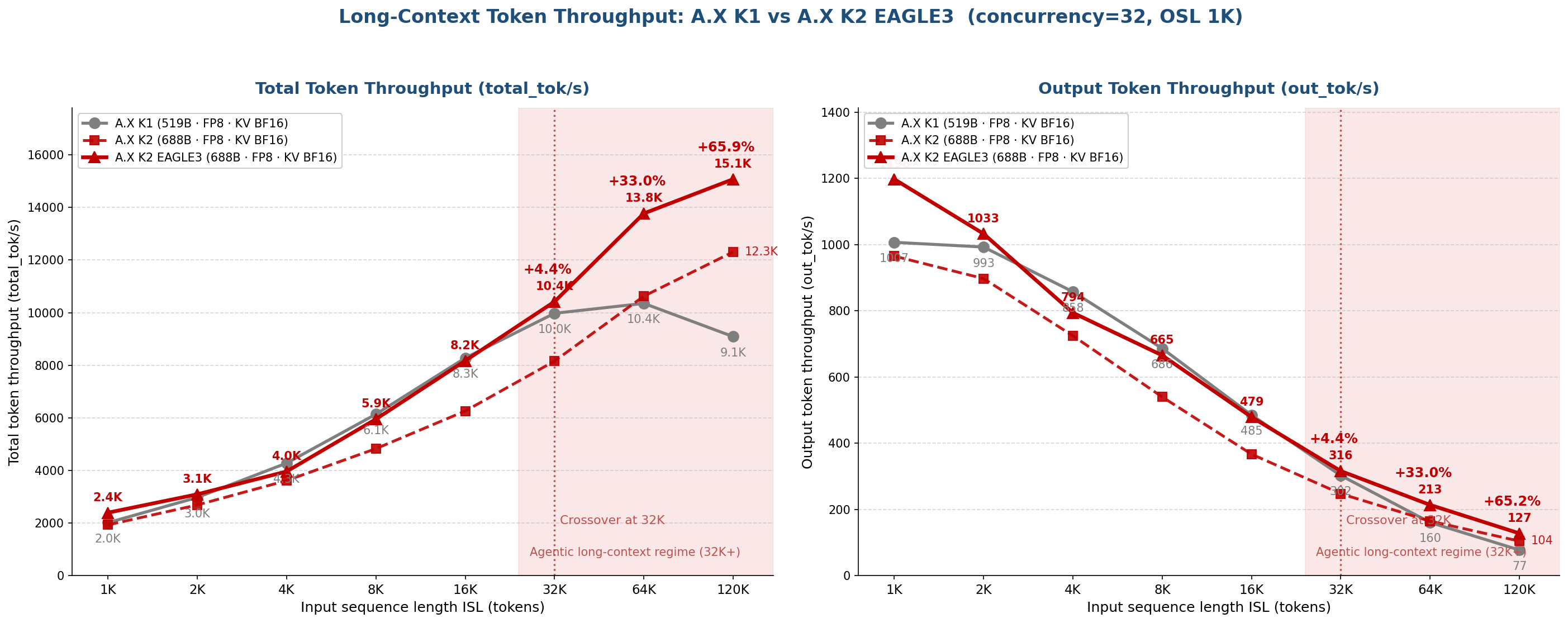}
    \caption{\textbf{Long-context serving throughput with EAGLE3 speculative decoding} (B200 single node, concurrency 32, OSL 1K). \textbf{Left:} total-token throughput; \textbf{right:} output-token throughput. EAGLE3 lifts A.X K2 by 23--30\% at every input length, advancing the crossover against A.X K1 from 64K to 32K and reaching $+65.9\%$ (15.1K total tok/s) at 120K.}
    \label{fig:throughput-eagle3}
\end{figure}

\paragraph{Decoding acceleration: EAGLE3 speculative decoding.} Speculative decoding with an EAGLE3 drafter lifts A.X K2's throughput by 23--30\% across all input lengths relative to the non-speculative configuration, moving the crossover against A.X K1 from 64K to 32K (Figure~\ref{fig:throughput-eagle3}): $+4.4\%$ at 32K (10.4K), $+33.0\%$ at 64K (13.8K), and $+65.9\%$ at 120K (15.1K). A single-layer MLA drafter proposes multiple tokens per step, and the A.X K2 target model verifies the proposals in one forward pass, accepting only the tokens that pass verification---on average 2.24 tokens are committed per step, cutting the number of decoding steps by more than half. The drafter checkpoint is 5.6\,GiB, under 1\% of the 688B target model, so it requires no additional GPUs or serving infrastructure: enabling it amounts to adding a single argument to the serving command.\footnote{Measured with vLLM \texttt{bench serve} on a random dataset. Speculative-decoding gains depend on the acceptance rate, i.e., how predictable the output distribution of the workload is; figures on real traffic such as code or dialogue may differ.}

\begin{figure}[t!]
    \centering
    \includegraphics[width=\textwidth]{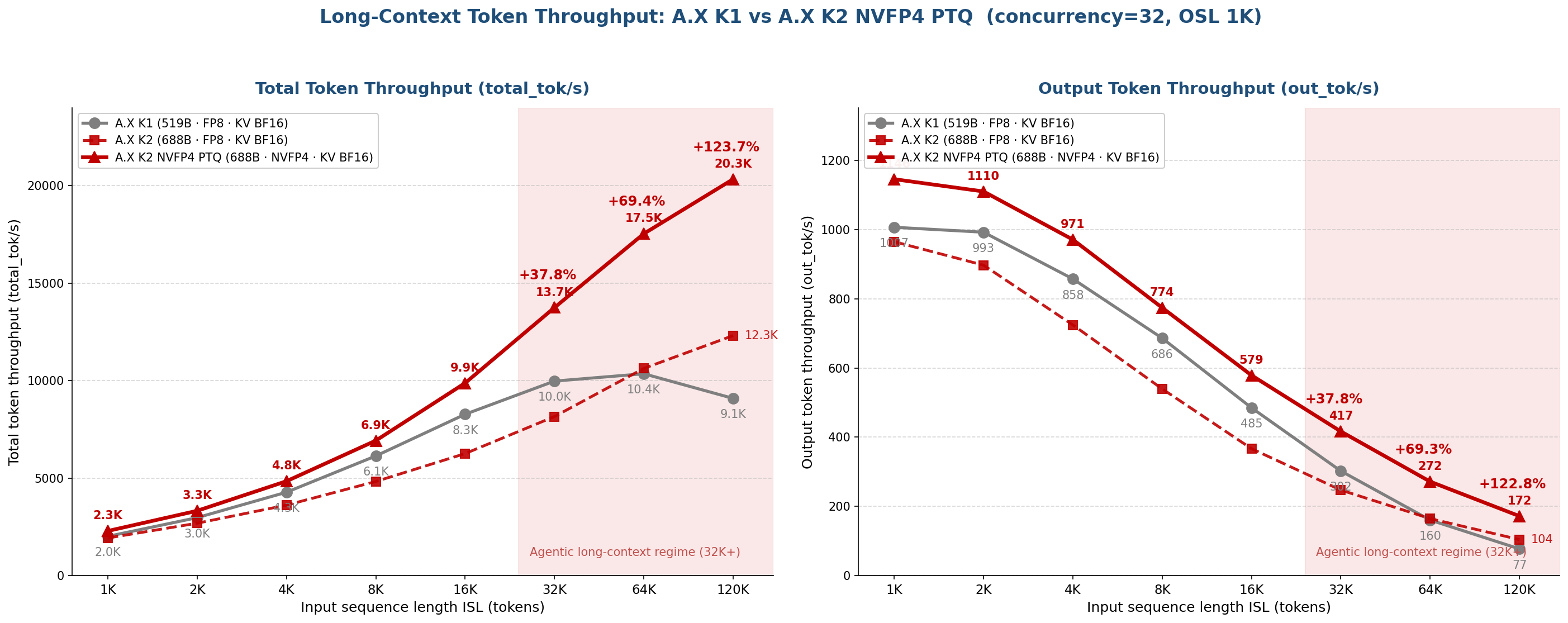}
    \caption{\textbf{Long-context serving throughput with NVFP4 PTQ} (B200 single node, concurrency 32, OSL 1K). \textbf{Left:} total-token throughput; \textbf{right:} output-token throughput. The NVFP4 configuration exceeds A.X K1 at every input length from 1K to 120K, reaching 20.3K total tok/s ($+123.7\%$) at 120K, with the margin widening in the agentic long-context regime (32K+).}
    \label{fig:throughput-nvfp4}
\end{figure}

\paragraph{Cost-optimized serving: NVFP4 PTQ.} NVFP4 4-bit post-training quantization improves throughput over the FP8 configuration at every input length, with the gains growing as the context lengthens; against A.X K1 it is ahead across the entire sweep, from 1K to 120K, and delivers roughly $1.4\times$--$2.2\times$ the throughput in the 32K+ long-context regime ($+37.8\%$ at 32K, $+69.4\%$ at 64K, $+123.7\%$ at 120K; Figure~\ref{fig:throughput-nvfp4}). Two factors matter here. First, halving the weight memory (FP8 $\rightarrow$ NVFP4) pairs with the native FP4 tensor-core acceleration of B200 Blackwell GPUs; since prefill computation dominates at long inputs, the FP4 compute advantage scales with context length. Second, the freed HBM converts directly into KV-cache headroom, keeping the configuration clear of the KV-capacity limit that the FP8 configuration approaches at very long inputs---extending the feasible operating envelope to higher concurrency and longer contexts. Because this comes at almost no accuracy cost (99.0\% of FP8 accuracy retained; Table~\ref{tab:fp8-nvfp4}), NVFP4 constitutes a practical operating configuration for throughput- and cost-prioritized workloads.

\paragraph{Summary.} The longer the context---coding, tool calling, accumulated histories, and other agentic use cases---the wider A.X K2's efficiency advantage becomes. Layering decoding acceleration (EAGLE3) and quantization (NVFP4) selectively on the SGA architectural base overturns the conventional expectation that a larger model must cost more to serve: with the optimizations applied, A.X K2 exceeds A.X K1 across the entire 32K+ agentic regime, and the NVFP4 option extends this to every input length. The 688B A.X K2 is thus simultaneously \emph{larger} and \emph{faster and cheaper to serve} than the 519B A.X K1, dissolving the trade-off between model scale and serving economics and establishing the technical basis for scaling to the next generation of large models.

\section{Limitations}
While A.X K2 demonstrates competitive performance and validates our efficient training and post-training recipe, several limitations remain, largely driven by real-world constraints and stability-focused engineering trade-offs. These limitations also delineate a clear roadmap for future iterations.

\begin{itemize}
    \item \textbf{Infrastructure Constraints on Scaling:} Our training environment imposed system-level constraints on parallelism. In particular, we confined expert parallelism to the eight-GPU intra-node NVLink domain (EP\,$=$\,8) to keep expert-parallel communication off slower inter-node links. This limited our exploration of higher-degree, multi-node expert parallelism, which could enable finer-grained expert partitioning; realizing it efficiently will require tighter integration between expert-parallel communication and the inter-node interconnect.

    \item \textbf{Optimization under Resource Constraints:} To maximize attainable performance under fixed time and GPU budget constraints, we selected model configurations guided by empirical scaling laws. Despite parameter-level optimization, these constraints led to modest performance shortfalls relative to similarly sized models trained with larger compute budgets. We expect that increasing available compute will allow us to relax these constraints and further improve model performance.

    \item \textbf{Scope of Modalities:} A.X K2 is currently text-only. While it exhibits strong reasoning performance, it lacks native multimodal understanding. We plan to extend the model with native multimodal capabilities in future work.
\end{itemize}

\section{Conclusion}
In this work, we presented A.X K2, a 688B-parameter MoE language model that bridges high-capacity reasoning with practical inference efficiency. Under fixed and time-bounded resources, we followed MoE scaling principles and used compute-budget-based planning to select a principled trade-off between model scale and training tokens, demonstrating that careful system-aware design can yield globally competitive capability.
We also introduced \textit{Think-Fusion}, a supervised fine-tuning recipe that trains on paired thinking and non-thinking responses, followed by multi-stage on-policy reinforcement learning, to enable explicit, user-controllable switching between \textit{thinking} and \textit{non-thinking} modes within a single unified model. Across mathematics, coding, and general-knowledge benchmarks in both English and Korean, our evaluations show that A.X K2 is competitive with strong open-weight baselines.

Beyond the compute-optimal architecture suggested by scaling laws, this project focuses on the system-level engineering required to translate that design into effective large-scale MoE training under real-world infrastructure constraints. Empirical profiling revealed systematic gaps between theoretical and realized throughput, driven by communication overhead, load imbalance, and execution-level inefficiencies. To preserve the intended training compute and schedule, we provisioned 512 NVIDIA B200 GPUs and iteratively optimized stability, utilization, and throughput over approximately 70 days of pretraining. This approach enabled stable convergence and predictable training behavior within fixed resource and timeline constraints.

Looking ahead, we will prioritize native multimodal capabilities and continued scaling toward the trillion-parameter regime. Building on the system and optimization insights established here, we aim to further advance both model capability and the efficiency of large-scale training and inference.

\appendix
\section{Contributors}

All authors are listed alphabetically by first name. Names marked with an asterisk (*) indicate authors whose affiliations differ from their affiliation at the time of this work.

\subsection{Model Engineering}
Cheolseung Baek, Eunki Kim, Hyunho Yang, Hyunjun Eun, Jin Kim, Junyoung Park, Juyun Wee, Minsang Kim, Minsoo Kang, Seongho Choi, Sangyeol Lee, Seokhwan Jo, Seonghye Cho*, Seongmin Ok, Subin Yi, Sung Jun Cheon, Sungwan Kim, Tae Yoon Kim

\subsection{Data Engineering}
Dhammiko Arya, Gun Song, Gyoungeun Han, Minki Hong*, Minkyung Park*, SaeRom Kim, Sangjin Kim, Seojin Lee, Seokyoung Hong, Sereimony Sek, Singon Kim*, Sohee Park, Sungbin Yoon, Sungeun Lee, Sunwoo Lee, Wonbeom Jang, Yohan Ra, Yong-jin Han, Yujin Kang*, Yujin Lee

\subsection{HPC Infrastructure}
Seungsik Kim, Youngrang Kim

\subsection{Business \& Compliance}
Seungmo Cho, Sooyeon Park, Youngjin Kim

\subsection*{Acknowledgment}
This work was supported by the Ministry of Science and ICT (MSIT), Republic of Korea, through the National IT Industry Promotion Agency (NIPA) (Grant No.~PJT-26-010018).
This research was also conducted as part of the Sovereign AI Foundation Model Project (Data Track), organized by the Ministry of Science and ICT (MSIT) and supported by the National Information Society Agency (NIA), Republic of Korea (Grant No.~2026-AIData-WII01).

\clearpage
\section{Appendix}

\subsection{Pre-train Data Category Mixture}\label{appendix:pretrain_data}

We report the designed composition of the pre-training mixture by language (Table~\ref{tab:dataset-language-stages}) and by content category (Table~\ref{tab:dataset-composition-stages}). Percentages are over each stage's designed mixture; the tokens actually consumed may be smaller (e.g., Stage~1 trained on the first 6.4T tokens of its $\sim$8T schedule, while Stage~2, Stage~3A, and Stages~3B\&3C used their full schedules of 1.4T, 0.22T, and 0.14T tokens, respectively). Stage~3B and Stage~3C share a single data mixture and are reported together. A dash (--) indicates a category not separately tracked for that stage, as the category taxonomy was refined across stages; the ``Etc.'' row aggregates remaining minor categories (including encyclopedia, news, and articles). The ``Academic Papers'' category is predominantly academic papers, with a small amount of other specialized long-form domain text such as legal documents and patents. The ``Instruction'' category consists of instruction-style (SFT-format) data introduced during pre-training, and the ``STEM'' category covers science and engineering domain text distinct from the academic-paper sources. Categories are grouped by data source/domain and by the target capability they primarily serve. Percentages are rounded to two decimals and may not sum exactly to 100\%.

\begin{table}[h]
\centering
\begin{tabular}{l r r r r}
\toprule
\textbf{Language} & \textbf{Stage 1 (\%)} & \textbf{Stage 2 (\%)} & \textbf{Stage 3A (\%)} & \textbf{Stage 3B\&3C (\%)} \\
\midrule
English   & 72.72  & 76.80  & 88.66  & 90.24 \\
Korean    & 15.40  & 14.33  & 7.07   & 6.15  \\
Code      & 8.29   & 7.35   & 3.54   & 2.99  \\
Japanese  & 1.41   & 0.50   & 0.24   & 0.20  \\
Spanish   & 1.15   & 0.52   & 0.25   & 0.21  \\
Chinese   & 1.02   & 0.51   & 0.25   & 0.21  \\
\midrule
Total     & 100.00 & 100.00 & 100.00 & 100.00 \\
\bottomrule
\end{tabular}
\caption{Language composition of the pre-training mixture across stages (\%).}
\label{tab:dataset-language-stages}
\end{table}

\begin{table}[h]
\centering
\begin{tabular}{l r r r r}
\toprule
\textbf{Category} & \textbf{Stage 1 (\%)} & \textbf{Stage 2 (\%)} & \textbf{Stage 3A (\%)} & \textbf{Stage 3B\&3C (\%)} \\
\midrule
\multicolumn{5}{@{}l}{\emph{By data source / domain}} \\
Web             & 69.91  & 25.81  & 12.43  & 10.49 \\
Code            & 8.56   & 18.83  & 9.07   & 7.65  \\
Academic Papers & 16.69  & 14.62  & 7.04   & 5.94  \\
STEM            & --     & 5.53   & 2.66   & 2.25  \\
Books           & 0.66   & 3.74   & 1.80   & 1.52  \\
Q\&A            & --     & 4.98   & 2.40   & 2.03  \\
\midrule
\multicolumn{5}{@{}l}{\emph{By target capability}} \\
Instruction     & 3.45   & 22.49  & 10.83  & 9.14  \\
Reasoning       & --     & 3.06   & 26.33  & 22.22 \\
Agent           & --     & --     & 15.14  & 25.55 \\
Long-context    & --     & --     & 11.85  & 12.82 \\
\midrule
Etc.            & 0.73   & 0.92   & 0.46   & 0.38  \\
\midrule
Total           & 100.00 & 100.00 & 100.00 & 100.00 \\
\bottomrule
\end{tabular}
\caption{Category composition of the pre-training mixture across stages (\%).}
\label{tab:dataset-composition-stages}
\end{table}

\subsection{SFT Data Statistics}\label{appendix:sft_data}
As mentioned in the main text, we periodically assessed the model's performance during training and supplemented the data accordingly. Table~\ref{tab:sft-token-ratio} and Table~\ref{tab:sft-length-dist} present the statistics over the entire set of samples used for SFT.

\begin{table}[htbp]
\centering
\small
\begin{tabular}{ll rr rr rr}
\toprule
 & & \multicolumn{2}{c}{Reasoning} & \multicolumn{2}{c}{Instruction} & \multicolumn{2}{c}{Total} \\
\cmidrule(lr){3-4}\cmidrule(lr){5-6}\cmidrule(lr){7-8}
Stage & Category & Tokens & Ratio (\%) & Tokens & Ratio (\%) & Tokens & Ratio (\%) \\
\midrule

\multirow{8}{*}{Indexer SFT}
 & Math                  & 1.19B & 1.96 & 19.45M & 0.03 & 1.21B & 1.99 \\
 & Knowledge / Science   & 0.86B & 1.41 & 3.44M & 0.01 & 0.86B & 1.42 \\
 & Instruction Following & 0.01B & 0.02 & 0.54M & 0.00 & 0.01B & 0.02 \\
 & Code                  & 0.54B & 0.89 & 0 & 0.00 & 0.54B & 0.89 \\
 & Agent                 & 1.32B & 2.18 & 91.63M & 0.15 & 1.41B & 2.33 \\
 & General               & 0.14B & 0.23 & 34.21M & 0.06 & 0.17B & 0.29 \\
 & Safety                & 0 & 0.00 & 1.97M & 0.00 & 1.97M & 0.00 \\
\cmidrule(l){2-8}
 & \textbf{Sum} & 4.06B & 6.70 & 151.2M & 0.25 & 4.21B & 6.95 \\
\midrule
\multirow{8}{*}{Main SFT}
 & Math                  & 15.00B & 24.76 & 0.61B & 1.01 & 15.61B & 25.77 \\
 & Knowledge / Science   & 11.21B & 18.50 & 0.11B & 0.18 & 11.32B & 18.68 \\
 & Instruction Following & 1.27B & 2.09 & 0.01B & 0.01 & 1.27B & 2.10 \\
 & Code                  & 6.79B & 11.20 & 0.09B & 0.15 & 6.88B & 11.35 \\
 & Agent                 & 14.68B & 24.24 & 1.13B & 1.86 & 15.81B & 26.10 \\
 & General               & 3.29B & 5.44 & 2.11B & 3.48 & 5.40B & 8.91 \\
 & Safety                & 0.06B & 0.10 & 0.02B & 0.04 & 0.08B & 0.14 \\
\cmidrule(l){2-8}
 & \textbf{Sum} & 52.30B & 86.32 & 4.08B & 6.73 & 56.38B & 93.05 \\
\midrule
\multicolumn{2}{l}{\textbf{Total}}
 & 56.36B & 93.02 & 4.23B & 6.98 & 60.59B & 100.00 \\
\bottomrule
\end{tabular}
\caption{Token composition and ratio (\%) across the SFT stages. Token counts are abbreviated as K (thousand), M (million), and B (billion).}
\label{tab:sft-token-ratio}
\end{table}


\begin{table}[htbp]
\centering
\small
\begin{tabular}{ll rr rr rr}
\toprule
 & & \multicolumn{2}{c}{Reasoning} & \multicolumn{2}{c}{Instruction} & \multicolumn{2}{c}{Total} \\
\cmidrule(lr){3-4}\cmidrule(lr){5-6}\cmidrule(lr){7-8}
Stage & Category & $<$32K & 32--128K & $<$32K & 32--128K & $<$32K & 32--128K \\
\midrule
\multirow{8}{*}{Indexer SFT}
 & Math                  & 0.37  & 0.15 & 0.03  & 0.00 & 0.40  & 0.15 \\
 & Knowledge / Science   & 0.89  & 0.09 & 0.05  & 0.00 & 0.95  & 0.09 \\
 & Instruction Following & 0.06  & 0.00 & 0.01  & 0.00 & 0.07  & 0.00 \\
 & Code                  & 0.38  & 0.06 & 0.00  & 0.00 & 0.38  & 0.06 \\
 & Agent                 & 1.02  & 0.15 & 0.02  & 0.01 & 1.03  & 0.16 \\
 & General               & 0.80  & 0.01 & 0.71  & 0.00 & 1.51  & 0.01 \\
 & Safety                & 0.00  & 0.00 & 0.04  & 0.00 & 0.04  & 0.00 \\
\cmidrule(l){2-8}
 & \textbf{Sum}          & 3.51  & 0.45 & 0.85  & 0.02 & 4.37  & 0.47 \\
\midrule
\multirow{8}{*}{Main SFT}
 & Math                  & 4.61  & 1.86 & 0.92  & 0.07 & 5.53  & 1.94 \\
 & Knowledge / Science   & 13.84 & 1.13 & 1.64  & 0.00 & 15.48 & 1.13 \\
 & Instruction Following & 5.53  & 0.00 & 0.09  & 0.00 & 5.62  & 0.00 \\
 & Code                  & 4.80  & 0.69 & 0.72  & 0.00 & 5.52  & 0.69 \\
 & Agent                 & 16.24 & 1.47 & 0.24  & 0.16 & 16.48 & 1.63 \\
 & General               & 19.20 & 0.15 & 20.90 & 0.02 & 40.10 & 0.16 \\
 & Safety                & 0.38  & 0.00 & 0.50  & 0.00 & 0.88  & 0.00 \\
\cmidrule(l){2-8}
 & \textbf{Sum}          & 64.60 & 5.31 & 25.00 & 0.25 & 89.60 & 5.56 \\
\midrule
\multicolumn{2}{l}{\textbf{Total}}
 & 68.11 & 5.76 & 25.86 & 0.27 & 93.97 & 6.03 \\
\bottomrule
\end{tabular}
\caption{Sample distribution by token length range, reported as ratio (\%), across the SFT stages. ``$<$32k'' and ``32--128k'' denote the proportion of samples in each token-length range.}
\label{tab:sft-length-dist}
\end{table}

\subsection{Safety Reasoning Data Construction}
\label{appendix:safety_reasoning_data}

The safety portion of the SFT mixture builds on ESSA~\citep{park2026essa}, which evolves compact safety specifications for domain--task pairs and uses them to synthesize deliberation-structured safety supervision. A.X K1 used the ESSA safety data as part of its post-training mixture. For A.X K2, we extend this recipe with an additional specification-evolution and data-synthesis pass tailored to the A.X K2 training format.

The construction pipeline consists of four stages. First, we collect English and Korean safety prompts covering harmful requests, safety-adjacent benign requests, and over-refusal-sensitive cases. Second, each prompt is assigned a domain and task label using the ESSA taxonomy. These labels are not treated as supervision targets; they are used only to choose the relevant safety specification \(S_{d,t}\) among the 50 evolved domain--task combinations. Third, a reasoning-capable teacher is prompted with the selected safety specification and the user request to produce a reasoning-mode answer. The prompt is designed so that the final answer does not cite internal specifications, rule identifiers, or taxonomy labels; instead, boundaries are explained in ordinary language and safe alternatives are provided when appropriate. Compared with the original ESSA synthesis format, which used explicit \texttt{\#\#thought}/\texttt{\#\#response} markers, the A.X K2 pipeline converts outputs into the standard chat-sample representation used by the target chat template.

Finally, we preserve raw request--response logs and source metadata before conversion, then run validation and filtering over the converted samples. The quality-control pass checks both safety and usefulness: unsafe operational residue, proxy harmful artifacts, privacy leakage, language mismatch, malformed reasoning/final-answer boundaries, and unnecessary refusal are routed to exclusion, regeneration, or meta-review. In particular, the filtering rubric rewards safe completion behavior: if the user's benign objective can be served without enabling harm, the preferred answer redirects to a useful safe artifact rather than issuing a blanket refusal.

\subsection{Details of Evaluation Method}
\label{subsec:details_of_evaluation_method}

\paragraph{Serving setup}
All open-weight baseline models are evaluated via OpenRouter, with the provider fixed to the model publisher only. We set the reasoning effort to \texttt{xhigh} and leave all generation parameters unset---using the provider's defaults---except for the maximum number of output tokens. As OpenRouter appears to cap the output at 64K tokens by default, we override it as follows: (1) where available, we use the maximum output token length stated on the OpenRouter model description page; (2) if the context limit is hit under this setting, we instead set the maximum output length to the stated maximum context length minus 100K tokens. For Qwen3.5-397B-A17B served by Alibaba, the maximum output length is fixed at 65,536 tokens. For all other models, the resulting maximum output length is at least 128K tokens, including when it is computed as the context length minus 100K tokens.

\paragraph{Metrics}
All results in Table~\ref{tab:LLM_large_latest_think} are reported as pass@1 scores averaged over multiple generations per sample, with the number of generations varying by benchmark. All mathematics benchmarks use eight generations per sample. The AA-Omniscience score follows the AAII v4.1 definition, $(2 \times \text{accuracy} + (1 - \text{hallucination rate})) / 3$, reported in percentages.

\paragraph{BrowseComp}
For BrowseComp, we use the LLM Context API from Brave Search as the only tool available to the model; no further web-page fetching or summarization is performed. Due to search cost, the number of searches is limited to at most 10 per problem.

\subsection{Details of Evaluation Benchmarks}
\label{subsec:details_of_evaluation_benchmarks}

Our evaluation prompt templates largely follow those of A.X~K1~\citep{skt2026axk1tr}; we summarize the benchmark suite here for completeness. For A.X K2, responses are generated for all benchmarks until the model reaches its context-length limit, and responses that hit the limit are always graded as incorrect.

\paragraph{Math}
We use AIME26~\citep{aime26}, a prestigious high-school–level mathematics competition problem set drawn from the 2026 American Invitational Mathematics Examination (AIME); Apex~\citep{dekoninck2026matharena}, a MathArena benchmark of 12 short-form-answer problems from 2025 mathematics competitions that then-state-of-the-art models could not solve, together with Apex-shortlist, a companion set of 48 less difficult problems on which those models reached roughly 50\% accuracy; and KMO26 (1st round), consisting of problems from the first round of the 2026 Korean Mathematical Olympiad. All math benchmarks are evaluated based on the correctness of the final answer, using a chat-based chain-of-thought (Chat CoT) prompting setup that encourages sufficient intermediate reasoning, and are reported as pass@1 averaged over 8 generations (Appendix~\ref{subsec:details_of_evaluation_method}). The evaluation prompt and a usage example for AIME26 are shown in Fig.~\ref{fig:aime26_chat}.

\begin{tcolorbox}[colback=gray!10,colframe=black,title=Evaluation Prompt for AIME26,fontupper=\small]
Solve the following math problem efficiently and clearly.  The last line of your response should be of the following format: 'Therefore, the final answer is: \$\textbackslash boxed\{ANSWER\}\$. I hope it is correct' (without quotes) where ANSWER is just the final number or expression that solves the problem. Think step by step before answering.\par
\vspace{\baselineskip}
\textcolor{gray}{Patrick started walking at a constant rate along a straight road from school to the park. One hour after Patrick left, Tanya started running along the same road from school to the park. One hour after Tanya left, Jose started bicycling along the same road from school to the park. Tanya ran at a constant rate of $2$ miles per hour faster than Patrick walked, Jose bicycled at a constant rate of $7$ miles per hour faster than Tanya ran, and all three arrived at the park at the same time. The distance from the school to the park is $\frac{m}{n}$ miles, where $m$ and $n$ are relatively prime positive integers. Find $m + n$.}
\end{tcolorbox}
\noindent\begin{minipage}{\textwidth}
\captionof{figure}{One example of evaluation prompt for AIME26. The model’s response is evaluated by comparing the \texttt{ANSWER} field against the ground-truth answer.}\label{fig:aime26_chat}
\end{minipage}

\paragraph{Korean}
For Korean evaluation, we use KMMLU-Pro~\citep{hong2025kmmluredux}, KoBALT~\citep{shin2025kobalt}, and CLIcK~\citep{kim2024click}. KMMLU-Pro includes questions based on the Korean national professional qualification exams, KoBALT measures Korean linguistic knowledge, and CLIcK evaluates Korean cultural and linguistic knowledge. CLIcK is evaluated in a standard zero-shot setting, consistent with the evaluation protocol used for pre-trained models.

\paragraph{Code}
We use LiveCodeBench v6~\citep{jain2024livecodebench}, SciCode~\citep{NEURIPS2024_36850592}, and Terminal Bench v2.1~\citep{tbench2026}. For LiveCodeBench, generated code is executed using a Python interpreter, and correctness is determined by matching execution results with expected outputs; since the LiveCodeBench series includes recently updated problems, reducing the risk of data contamination, we evaluate only the February--May 2025 subset. SciCode consists of research-level scientific programming problems curated by scientists. Terminal Bench evaluates the ability to complete realistic tasks in a terminal environment; its scores are taken from Artificial Analysis.

\paragraph{Science \& Knowledge}
We use GPQA Diamond~\citep{rein2024gpqa}, Humanity's Last Exam~\citep{hle2026nature}, and AA-Omniscience~\citep{artificialanalysis2025omniscience}. GPQA Diamond consists of high-difficulty, graduate-level questions in physics, chemistry, and biology, and HLE evaluates advanced knowledge and reasoning capabilities. AA-Omniscience jointly measures factual knowledge and hallucination propensity; we report the AAII v4.1 score from Artificial Analysis, defined in Appendix~\ref{subsec:details_of_evaluation_method}.

\paragraph{General}
We use IFBench~\citep{pyatkin2025generalizing}, which assesses compliance with explicit constraints such as output format, length limits, required expressions, or language usage under rule-based criteria, and AA-LCR~\citep{artificialanalysis2025lcr}, which evaluates the model's ability to maintain coherence and retain key information when generating responses based on long input contexts.

\paragraph{Agentic}
GDPval~\citep{openai2025gdpval} evaluates performance on real-world, economically valuable professional tasks; we report the Elo rating from the Artificial Analysis GDPval v2 arena. $\tau^2$-Bench~\citep{barres2025tau2}, building on $\tau$-bench~\citep{yao2024tau}, and its successor $\tau^3$-Bench~\citep{shi2026tau} evaluate conversational agents in a dual-control environment; we report the Telecom domain score of $\tau^2$-Bench and the Banking domain score of $\tau^3$-Bench, both from Artificial Analysis. BrowseComp~\citep{wei2025browsecomp} evaluates browsing agents on locating hard-to-find information on the web; our search-tool configuration is described in Appendix~\ref{subsec:details_of_evaluation_method}.

\subsection{Multilingual Evaluation}
\label{appendix:globalmmlu}

The benchmark suite in Appendix~\ref{subsec:details_of_evaluation_benchmarks} measures Korean and English capability but does not isolate multilingual knowledge. We therefore evaluate the multilingual capability of A.X K2 on Global-MMLU-Lite~\citep{singh2025globalmmlu}, a subset of MMLU~\citep{hendryckstest2021} whose items have been human-translated and post-edited so that they are parallel across languages. Since Korean is already covered by the Korean benchmarks, this evaluation uses the four non-Korean languages, with 400 four-way multiple-choice items per language in English, Spanish, Chinese, and Japanese, for a total of 1,600 items. All models are evaluated in thinking mode, and all items are evaluated 0-shot: the model is asked to output only the option symbol, and the response is scored by exact match. Generation uses \texttt{temperature}\,$=0$ with a single generation per item. As shown in Table~\ref{tab:globalmmlu}, A.X K2 improves over A.X K1 in all four languages, and the improvement is concentrated in the non-English languages.

\begin{table}[htbp]
\centering
\small
\begin{tabular}{l rrrrr}
\toprule
\textbf{Model} & \textbf{Avg} & \textbf{English} & \textbf{Spanish} & \textbf{Chinese} & \textbf{Japanese} \\
\midrule
A.X K2              & 87.56 & 89.75 & 87.75 & 87.00 & 85.75 \\
A.X K1              & 77.75 & 89.50 & 81.25 & 72.25 & 68.00 \\
\midrule
Qwen3.5-397B-A17B   & \textbf{91.88} & \textbf{93.50} & \textbf{92.50} & 90.50 & \textbf{91.00} \\
Nemotron 3 Ultra    & 90.94 & \textbf{93.50} & 90.25 & \textbf{90.75} & 89.25 \\
DeepSeek-V4 Flash   & 86.81 & 90.25 & 88.75 & 84.00 & 84.25 \\
GLM-5.1             & 90.31 & 91.75 & 90.75 & 89.50 & 89.25 \\
Kimi-K2.6           & 90.62 & 92.50 & 91.00 & 89.50 & 89.50 \\
MiniMax-M2.7        & 87.31 & 90.00 & 89.00 & 83.75 & 86.50 \\
\bottomrule
\end{tabular}
\caption{\textbf{Multilingual performance on Global-MMLU-Lite in thinking mode} (exact match, \%). The highest model score in each column is shown in bold, including ties.}
\label{tab:globalmmlu}
\end{table}

A.X K2 attains 87.56, ahead of DeepSeek-V4 Flash (86.81) and MiniMax-M2.7 (87.31) while trailing the strongest baselines. Compared with A.X K1, the most notable change is that the average score improves while the variation across languages decreases substantially. The range across languages---the highest language score minus the lowest---falls from 21.50 points for A.X K1 (English 89.50, Japanese 68.00) to 4.00 points for A.X K2 (English 89.75, Japanese 85.75), approaching the most language-consistent baselines, GLM-5.1 (2.50 points) and Qwen3.5 (3.00 points). These results indicate that A.X K2 varies less with the language of the query than A.X K1, and responds more consistently to knowledge items presented in different languages.

\subsection{LongBench v2: Full-Set and Category-Level Results}
\label{appendix:longbenchv2-full}

Section~\ref{subsec:long_context_eval} reports LongBench v2 results after removing the 87 items that contain Han characters. Table~\ref{tab:longbench-v2-full} reports results on all 503 items to show how that filter affects the comparison. Each model is evaluated once with the official harness, and the 416-item scores in Table~\ref{tab:longbench-v2} are recomputed from the same predictions after filtering, so only the set of scored items differs between the two tables.

\begin{table}[htbp]
\centering
\small
\begin{tabular}{l *{6}{c}}
\toprule
\textbf{Model} & \textbf{Overall} & \textbf{Easy} & \textbf{Hard} & \textbf{Short} & \textbf{Medium} & \textbf{Long} \\
\midrule
Qwen3.5-397B-A17B  & \textbf{64.6} & \textbf{72.9} & 59.5          & 67.2          & \textbf{65.1} & 59.3          \\
GLM-5.1            & 63.8          & 69.8          & \textbf{60.1} & \textbf{70.6} & 59.5          & \textbf{61.1} \\
A.X K2             & 62.2          & 70.8          & 56.9          & 67.2          & 60.0          & 58.3          \\
Nemotron 3 Ultra   & 61.4          & 69.3          & 56.6          & 65.0          & 61.9          & 54.6          \\
\midrule
Human expert       & 53.7          & --            & --            & 47.2          & 59.1          & 53.7          \\
\bottomrule
\end{tabular}
\caption{\textbf{LongBench v2 accuracy (\%) on the full 503-item set}, i.e.\ including the 87 items that contain Han characters. The set splits into Easy 192 / Hard 311 by difficulty and Short 180 / Medium 215 / Long 108 by input length. Rows are ordered by overall score. The highest model score in each column is shown in bold, including ties; the human-expert baseline is not included in this comparison. All scores are measured with the official LongBench v2 harness under the protocol of Table~\ref{tab:longbench-v2}: baselines are served through OpenRouter as described in Appendix~\ref{subsec:details_of_evaluation_method}, and A.X K2 is evaluated as released, with SGA sparse attention and FP8 weights, using a 200K input budget under YaRN scaling with a scaling factor of 2 (Sec.~\ref{subsec:long_context}). The human-expert row is the official baseline of \citet{bai2025longbenchv2}, collected under a 15-minute time limit per item; the authors do not report it by difficulty.}
\label{tab:longbench-v2-full}
\end{table}

\paragraph{Effect of the filter across models.} Comparing Table~\ref{tab:longbench-v2-full} with Table~\ref{tab:longbench-v2}, A.X K2 is the only model whose overall score increases after filtering, from 62.2 to 63.9 ($+1.7$). The scores of GLM-5.1, Qwen3.5, and Nemotron 3 Ultra decrease by 0.6, 0.4, and 0.1 points, respectively. Consequently, A.X K2 ranks third on the full set but second on the filtered set, 0.3 points behind Qwen3.5. The result on the \textbf{Long} split also changes: GLM-5.1 leads on the full set with 61.1, followed by Qwen3.5 at 59.3 and A.X K2 at 58.3, whereas all three score 56.2 after filtering. We report the filtered set in the main text because it better matches the benchmark's stated English-only design. We report the full set as well because the filter affects models unevenly.

\paragraph{Slice-level effect on A.X K2.} Table~\ref{tab:longbench-v2-delta} compares A.X K2's scores before and after filtering. Although the overall score increases by 1.7 points, the \textbf{Long} score decreases by 2.1 points. The largest decreases occur in Code Repository Understanding ($-14.0$) and Long Structured Data Understanding ($-10.6$), for which the filter removes 40\% and 33\% of the items, respectively. Long-dialogue History Understanding is unchanged because none of its items are removed.

\begin{table}[htbp]
\centering
\small
\begin{tabular}{l rr rr r}
\toprule
& \multicolumn{2}{c}{\textbf{Full set}} & \multicolumn{2}{c}{\textbf{Filtered set}} & \\
\cmidrule(lr){2-3} \cmidrule(lr){4-5}
\textbf{Slice} & $n$ & \textbf{Score} & $n$ & \textbf{Score} & $\Delta$ \\
\midrule
Overall & 503 & 62.2 & 416 & 63.9 & $+1.7$ \\
\midrule
\multicolumn{6}{l}{\emph{By difficulty}} \\
Easy   & 192 & 70.8 & 150 & 74.7 & $+3.9$ \\
Hard   & 311 & 56.9 & 266 & 57.9 & $+1.0$ \\
\midrule
\multicolumn{6}{l}{\emph{By input length}} \\
Short  & 180 & 67.2 & 156 & 69.2 & $+2.0$ \\
Medium & 215 & 60.0 & 180 & 62.8 & $+2.8$ \\
Long   & 108 & 58.3 &  80 & 56.2 & $-2.1$ \\
\midrule
\multicolumn{6}{l}{\emph{By task category}} \\
Single-Document QA                  & 175 & 62.3 & 149 & 65.8 & $+3.5$ \\
Multi-Document QA                   & 125 & 60.8 & 108 & 63.9 & $+3.1$ \\
Long In-context Learning            &  81 & 71.6 &  68 & 75.0 & $+3.4$ \\
Code Repository Understanding       &  50 & 54.0 &  30 & 40.0 & $-14.0$ \\
Long-dialogue History Understanding &  39 & 74.4 &  39 & 74.4 & $0.0$ \\
Long Structured Data Understanding  &  33 & 42.4 &  22 & 31.8 & $-10.6$ \\
\bottomrule
\end{tabular}
\caption{\textbf{Effect of the Han-character filter on A.X K2}, measured on the same set of predictions. The full set is all 503 items and the filtered set is the 416 items that contain no Han characters; $\Delta$ is the change in accuracy (percentage points). The filter moves the score in both directions across slices.}
\label{tab:longbench-v2-delta}
\end{table}

\paragraph{Task-category breakdown.} Table~\ref{tab:longbench-v2-category} gives the category-level results for both item sets. A.X K2 has the highest Multi-Document QA score on both sets (60.8 and 63.9), ties for the highest Single-Document QA score on the filtered set (65.8), and has the highest Long In-context Learning score on the filtered set (75.0). It trails the leading models in Code Repository Understanding and Long Structured Data Understanding on both sets. These are also the smallest categories after filtering, with 30 and 22 items, so one additional correct answer changes the score by 3.3 and 4.5 points, respectively. The category results should be interpreted as broad indicators rather than precise measurements of each gap.

\begin{table}[htbp]
\centering
\small
\setlength{\tabcolsep}{4pt}
\begin{tabular}{l *{6}{c}}
\toprule
\textbf{Model} & \makecell{\textbf{Single-Doc} \\ \textbf{QA}} & \makecell{\textbf{Multi-Doc} \\ \textbf{QA}} & \makecell{\textbf{Long} \\ \textbf{ICL}} & \makecell{\textbf{Code} \\ \textbf{Repo}} & \makecell{\textbf{Long} \\ \textbf{Dialogue}} & \makecell{\textbf{Long} \\ \textbf{Struct.}} \\
\midrule
\multicolumn{7}{@{}l}{\emph{Full set} ($n=503$)} \\
\quad $n$          & 175           & 125           & 81            & 50            & 39            & 33            \\
Qwen3.5-397B-A17B  & 62.3          & 58.4          & \textbf{75.3} & \textbf{64.0} & 76.9          & 60.6          \\
GLM-5.1            & \textbf{63.4} & 57.6          & 72.8          & \textbf{64.0} & 74.4          & 54.5          \\
A.X K2             & 62.3          & \textbf{60.8} & 71.6          & 54.0          & 74.4          & 42.4          \\
Nemotron 3 Ultra   & 60.6          & 53.6          & 70.4          & 54.0          & \textbf{79.5} & \textbf{63.6} \\
\midrule
\multicolumn{7}{@{}l}{\emph{Filtered set} ($n=416$)} \\
\quad $n$          & 149           & 108           & 68            & 30            & 39            & 22            \\
Qwen3.5-397B-A17B  & 63.1          & 59.3          & 72.1          & \textbf{60.0} & 76.9          & \textbf{54.5} \\
GLM-5.1            & \textbf{65.8} & 59.3          & 69.1          & 53.3          & 74.4          & 40.9          \\
A.X K2             & \textbf{65.8} & \textbf{63.9} & \textbf{75.0} & 40.0          & 74.4          & 31.8          \\
Nemotron 3 Ultra   & 60.4          & 56.5          & 66.2          & 53.3          & \textbf{79.5} & \textbf{54.5} \\
\bottomrule
\end{tabular}
\caption{\textbf{LongBench v2 accuracy (\%) by task category}, on the full 503-item set and on the 416-item filtered set. Column headings abbreviate the official categories: Single-Document QA, Multi-Document QA, Long In-context Learning, Code Repository Understanding, Long-dialogue History Understanding, and Long Structured Data Understanding. The $n$ rows give the number of items per category in each set; the filter removes 40\% of Code Repository Understanding items and 33\% of Long Structured Data Understanding items but none of Long-dialogue History Understanding, whose scores are identical by construction. The highest model score in each column is shown in bold within each panel, including ties.}
\label{tab:longbench-v2-category}
\end{table}

\clearpage
\bibliography{references}
\bibliographystyle{colm2024_conference}

\end{document}